\documentclass[runningheads]{llncs}

\usepackage{eccv}

\usepackage{eccvabbrv}

\usepackage{graphicx}
\usepackage{booktabs}
\usepackage{threeparttable}
\usepackage{multirow}
\usepackage{wrapfig}
\usepackage{array}

\usepackage[accsupp]{axessibility}  

\usepackage[pagebackref,breaklinks,colorlinks,citecolor={green!70!black}]{hyperref}

\usepackage{orcidlink}

\begin{document}

\title{Visual Grounding for Object-Level Generalization in Reinforcement Learning}

\titlerunning{COPL}

\author{Haobin Jiang\inst{1} \and Zongqing Lu\inst{1,2}\thanks{Corresponding author.}}

\authorrunning{H.~Jiang and Z.~Lu}

\institute{School of Computer Science, Peking University \and
Beijing Academy of Artificial Intelligence \\
\texttt{\{haobin.jiang,zongqing.lu\}@pku.edu.cn}}

\maketitle

\begin{abstract}
  Generalization is a pivotal challenge for agents following natural language instructions. To approach this goal, we leverage a vision-language model (VLM) for visual grounding and transfer its vision-language knowledge into reinforcement learning (RL) for object-centric tasks, which makes the agent capable of zero-shot generalization to unseen objects and instructions. By visual grounding, we obtain an object-grounded confidence map for the target object indicated in the instruction. Based on this map, we introduce two routes to transfer VLM knowledge into RL. Firstly, we propose an object-grounded intrinsic reward function derived from the confidence map to more effectively guide the agent towards the target object. Secondly, the confidence map offers a more unified, accessible task representation for the agent's policy, compared to language embeddings. This enables the agent to process unseen objects and instructions through comprehensible visual confidence maps, facilitating zero-shot object-level generalization. Single-task experiments prove that our intrinsic reward significantly improves performance on challenging skill learning. In multi-task experiments, through testing on tasks beyond the training set, we show that the agent, when provided with the confidence map as the task representation, possesses better generalization capabilities than language-based conditioning. The code is available at \href{https://github.com/PKU-RL/COPL}{https://github.com/PKU-RL/COPL}.
  \keywords{Reinforcement learning \and Visual grounding}
\end{abstract}

\section{Introduction}
\label{sec:intro}

In the field of artificial intelligence, the ability of agents to understand and follow natural language instructions in an open-ended manner is crucial \cite{brohan2022rt,brohan2023rt,chen2023open,shah2023lm}. However, the scope of training content for an agent's policy learning is always finite. Zero-shot generalization task which involves being instructed to interact with diverse objects not encountered during training, from the vast realm of human vocabulary, represents a pivotal step towards creating general artificial intelligence systems capable of adapting to a wide range of real-world scenarios \cite{chen2023open,stone2023open}. As a popular open-ended 3D game, Minecraft serves as an ideal testbed for learning and evaluating generalization ability. At its core, Minecraft offers procedurally generated worlds with unlimited size and a large variety of tasks ranging from navigation and combat to building and survival \cite{fan2022minedojo,wang2023describe,yuan2023plan4mc,wang2023voyager,zhu2023ghost}. Compared with canonical game environments such as Go \cite{silver2016mastering}, Atari \cite{mnih2013playing}, and StarCraft \cite{vinyals2019grandmaster,ellis2024smacv2}, Minecraft mirrors the complexity of real-world challenges and offers a wide range of objects and tasks with natural language instructions.

\begin{figure}[t]
    \centering
    \vspace{4mm}
    \includegraphics[width=0.44\columnwidth]{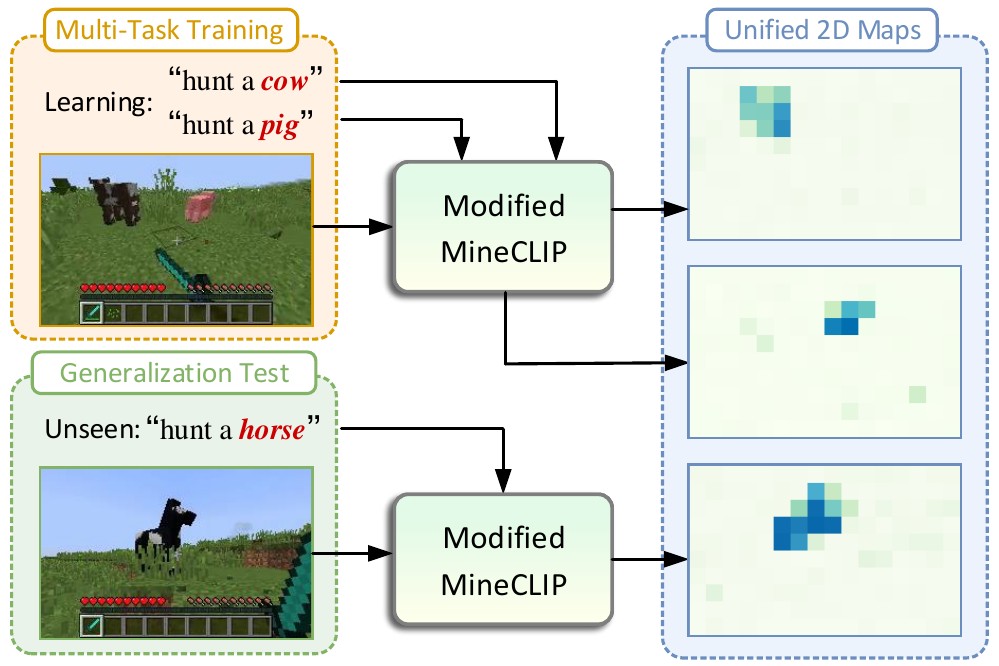}
    \includegraphics[width=0.54\columnwidth]{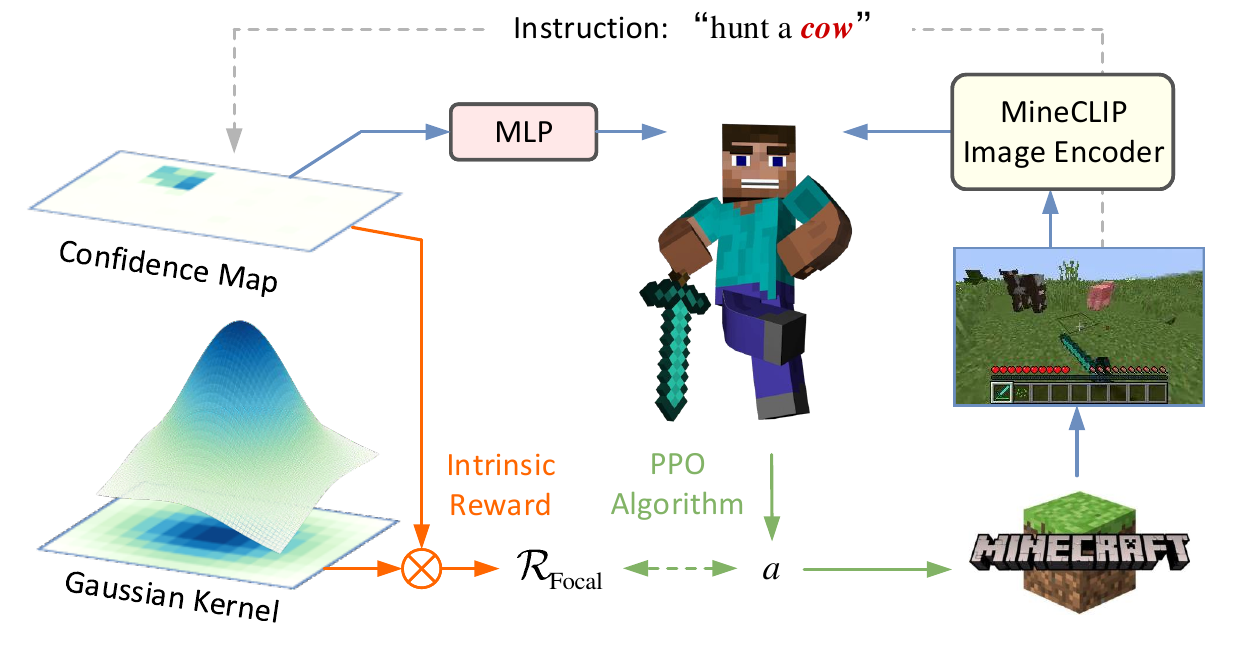}
    \vspace{4mm}
    \caption{Overview of \textbf{C}LIP-guided \textbf{O}bject-grounded \textbf{P}olicy \textbf{L}earning (\textbf{COPL}). 
    (\textit{left}) \textbf{Visual grounding}: The instruction is converted into a unified 2D confidence map via our modified MineCLIP.
    (\textit{right}) \textbf{Transfer VLM knowledge into RL}: The agent takes the confidence map as the task representation and is trained with our proposed focal reward derived from the confidence map to guide the agent toward the target.
    }
    \label{fig:illu}
    \vspace{-1mm}
\end{figure}

Given the finite data scope for the agent's policy learning, it is infeasible for the policy to directly comprehend the vast array of object names beyond the training set, which human language instructions might contain. To equip an agent with zero-shot generalization ability over objects, integration of a vision-language model (VLM) is a promising way \cite{wu2023towards}. A VLM aligns images and language vocabularies into the same feature space, bridging the gap between visual observations and natural language instructions. Therefore, it has the capability to ground the agent's unseen text, \textit{e.g.}, names of novel objects, in visual observations, enabling the agent to comprehend instructions not encountered during training. CLIP \cite{radford2021learning} emerges as a significant model and has become widely used \cite{nichol2022glide,ramesh2022hierarchical,shen2021much,shridhar2022cliport,gadre2023cows,minderer2205simple}. Recent works have extensively adopted CLIP as a foundation model for open-vocabulary object detection \cite{gu2021open,kuo2022open,zang2022open} and open-vocabulary segmentation \cite{ding2022open,rao2022denseclip,liang2023open}, leveraging its rich vision-language knowledge. Moreover, CLIP even exhibits remarkable segmentation capabilities and explainability without fine-tuning \cite{zhou2022extract,li2023clip}.

CLIP's visual grounding ability to perform segmentation inspires two routes for enhancing the agent's policy learning in Minecraft by transferring VLM knowledge into reinforcement learning (RL). The first approach is \emph{transfer via reward}. The pixel area of the target object can be used as a surrogate for the distance between the agent and the target object. It can then serve as an intrinsic reward \cite{aubret2019survey} to guide the agent towards the target object, thereby facilitating interaction. The second one is \emph{transfer via representation}. The segmentation result can replace the language instruction as a unified, more accessible task representation. Practical research in robotics proves that models with such location input show superior performance compared to mere text input \cite{stone2023open}. Most importantly, the segmentation is open-vocabulary \cite{radford2021learning,zhou2022extract}, which means it remains effective for instructions containing novel objects not encountered during agent training.

Thanks to MineCLIP \cite{fan2022minedojo}, a variant of CLIP fine-tuned on Internet-scale Minecraft videos from YouTube, it becomes accessible to develop a generalizable agent in Minecraft, following the aforementioned inspirations. Initially, MineCLIP is merely used as a tool to measure the similarity between a sequence of visual observations and the instruction. It serves as an intrinsic reward for RL and achieves notable performance in Minecraft. Based on this foundation, our goal is to further explore the potential capabilities of MineCLIP, enabling it to offer additional visual grounding information beyond the similarity between observations and instructions to aid the agent's policy learning and improve its generalization ability.

In this paper, we propose a \textbf{C}LIP-guided \textbf{O}bject-grounded \textbf{P}olicy \textbf{L}earning method, namely \textbf{COPL}, that transfers the vision-language knowledge contained in MineCLIP to RL at a minimal cost. By visual grounding, we generate a confidence map of the target object indicated in the language instruction via our modified MineCLIP. We extend MineCLIP with modifications inspired by MaskCLIP \cite{zhou2022extract} so that it can segment the specified object from the image. As illustrated in \Cref{fig:illu} (\textit{left}), our approach can convert instructions into unified two-dimensional confidence maps. To leverage this object-grounded result, we first design an intrinsic reward that takes into account the pixel area and location of the target object in the image observation. By doing so, we also address a deficiency of the original MineCLIP reward \cite{fan2022minedojo}: it is insensitive to the distance to the target object \cite{radford2021learning,cai2023open}. Furthermore, we integrate the resulting confidence map into the policy input as a task representation, as illustrated in \Cref{fig:illu} (\textit{right}). Based on this adjustment, our agent can possess zero-shot generalization capability over objects through multi-task RL trained on \textit{only a limited set of instructions.}
Due to the reuse of MineCLIP's encoding, such generalization capability does not incur significant additional computational overhead.

We evaluate COPL on basic skill learning and zero-shot object-level generalization in Minecraft. Firstly, we conduct a group of single-task experiments to show that our object-grounded intrinsic reward successfully enables the agent to acquire various challenging basic skills while the MineCLIP reward fails \cite{fan2022minedojo,cai2023open}. Then we extend our evaluation to instruction-following scenarios, where we train the agent with a set of instructions. In our test, the agent exhibits the capacity to execute instructions involving previously unseen targets, effectively demonstrating its generalization ability over objects. 
COPL's success rate on unseen targets surpasses that of language-conditioned methods by around 300\% in the hunting domain and 100\% in the harvest domain. 
Though we implement and evaluate COPL in Minecraft, we believe our method is extendable to other similar open-ended environments and draws insights into transferring VLM knowledge into RL for training generalizable agents.

\section{Preliminary}
\label{sec:pre}

\textbf{Problem Statement.} In this paper, we focus on \emph{object-centric} tasks in Minecraft, where the agent is instructed to interact with diverse objects. By zero-shot \emph{object-level} generalization, we mean that the agent is instructed to interact with objects beyond the training scope without fine-tuning during evaluation. To formalize, we denote the set of objects with which the agent learns to interact during the training phase as $C_t$, and the set of objects with which the agent is required to interact during the evaluation phase as $C_e$. To test the generalization ability of the agent, $C_e$ consists of objects that are not in $C_t$. For example, during training, the agent learns to accomplish language instructions ``\texttt{hunt a cow}'' and ``\texttt{hunt a sheep}''. However, during evaluation, it will encounter instructions like ``\texttt{hunt a horse}'' or ``\texttt{hunt a chicken}'', where neither \texttt{horse} nor \texttt{chicken} appears in the instructions during training. Note that we do not consider generalization ability concerning unseen
 actions, which could be left as future work. Therefore, instructions during evaluation should have the \emph{same behavior patterns} as those learned in training. For instance, when training with ``\texttt{hunt sth.}'' and ``\texttt{harvest sth.}'', testing with ``\texttt{explore the world}'' is not considered.

\vspace{1mm}
\noindent \textbf{Zero-Shot Generalization.} Our zero-shot object-level generalization in\\ Minecraft is a specific form of broader zero-shot generalization (ZSG) defined in the contextual Markov decision process (CMDP) framework \cite{kirk2023survey} in RL. While ZSG typically involves adapting to new environments or tasks, our focus is on enabling agents to follow instructions and interact with novel objects not encountered during training. This object-level adaptation aligns with ZSG's objective of performing effectively in unseen scenarios. In other words, the ``context'' in our case is specifically defined as the target objects indicated in instructions.

\vspace{1mm}
\noindent \textbf{MineCLIP for Minecraft RL.} MineCLIP is a VLM pre-trained on Internet-scale Minecraft videos from YouTube \cite{fan2022minedojo}, learning the alignment between video clips (16 frames) and natural language. Similar to CLIP \cite{radford2021learning}, MineCLIP adopts a ViT \cite{dosovitskiy2020image} as the image encoder and a GPT \cite{radford2019language} as the text encoder. The main difference between MineCLIP and CLIP is that MineCLIP takes as input a sequence of 16 images. Therefore, MineCLIP incorporates an additional module to aggregate the 16 embeddings generated by the image encoder. The proposed two mechanisms include a temporal transformer (MineCLIP[attn]) and direct average pooling (MineCLIP[avg]). In this paper, we choose the former as our base model due to its better performance in Programmatic tasks compared to the latter \cite{fan2022minedojo}. For RL in Minecraft, MineCLIP provides an intrinsic reward function $\mathcal{R}_i:\mathcal{G}\times \mathcal{S}^{16}\rightarrow \mathbb{R}$, representing the similarity between the image observation sequence of the previous 16 steps $[s_{t-15},\cdots,s_{t-1},s_t]$ and the task prompt $g$.

\section{Related Work}

\textbf{Minecraft Research.} 
Broadly, challenges in Minecraft can be categorized into high-level task planning and low-level skill learning. For high-level planning, where agents must make decisions on which skills to employ sequentially based on the given instruction, the field has converged towards leveraging large language models (LLMs) \cite{nottingham2023embodied,wang2023describe,wang2023voyager,yuan2023plan4mc,zhu2023ghost,wang2023jarvis}. Regarding learning low-level skills, the difficulty lies in the absence of well-defined dense reward and a vast variety of objects to interact with in Minecraft. Unlike the convergence in high-level planning approaches, two distinct routes have emerged in low-level learning. The first route, represented by MineCLIP \cite{fan2022minedojo}, utilizes the reward derived from the alignment between text and video clip or other manually designed rewards for RL \cite{yuan2023plan4mc}. The second one follows the principles of VPT \cite{baker2022video}, where skills are acquired through imitation learning based on large-scale demonstration \cite{cai2023open,lifshitz2023steve,cai2023groot}. Our work falls in the scope of low-level skill learning with RL.

\vspace{1mm}
\noindent \textbf{Instruction-Following RL.} Language has been widely explored in goal-\\conditioned RL for its compositional structure \cite{luketina2019survey}. This feature allows goal-conditioned policies to better capture the latent structure of the task space and generalize to unseen instructions that combine seen words \cite{oh2017zero,chan2019actrce,jiang2019language,colas2020language,mirchandani2021ella}. With the development of LLM and VLM, language also becomes a means of providing intrinsic rewards in RL. The similarity or correlation between instructions and current states provides dense rewards to guide the agent's learning more effectively \cite{fan2022minedojo,kwon2022reward,mahmoudieh2022zero,du2023guiding}. Our work stands out by enabling the policy to generalize to instructions that contain previously unseen targets.

\vspace{1mm}
\noindent \textbf{CLIP for Embodied AI.} CLIP \cite{radford2021learning} provides diverse usage for AI research. We categorize these applications into three areas: \emph{encoding}, \emph{retrieving} and \emph{locating}. Encoding, the most common use of CLIP, leverages CLIP encoders to represent images and/or texts \cite{shridhar2022cliport,khandelwal2022simple,majumdar2022zson}. Our work also utilizes the MineCLIP image encoder to process raw image observations. Retrieving mostly involves navigation tasks, where CLIP assists in selecting the most matching image from a set based on the given instruction \cite{dorbala2022clip,bucker2023latte,chen2023open,shah2023lm}. The most relevant usage to our work is locating, which applies methods like MaskCLIP \cite{zhou2022extract} or GradCAM \cite{selvaraju2017grad} on CLIP to determine the position of the specific object in images \cite{wang2022programmatically,gadre2023cows,zhang2023universal}. Based on the object location, agents can conduct planning with a depth detector \cite{gadre2023cows} or imitation learning \cite{wang2022programmatically,zhang2023universal}. In contrast, our work focuses on training agents via RL with information solely extracted from image observations, without any extra spatial information or demonstration.

\section{Method}
\label{sec:method}

In this section, we detail the implementation of our COPL method in Minecraft. We introduce how to exploit the visual grounding capability of MineCLIP through modifications, enabling the segmentation of the target object indicated in the language instruction (\Cref{sec:seg}). This process yields an object-grounded confidence map, where each element represents the probability of the specified target's presence. Based on this confidence map, we first implement VLM knowledge transfer via reward, presenting a simple but effective object-grounded intrinsic reward to guide the agent toward the target (\Cref{sec:reward}). Then, we propose transfer via representation, where we integrate the confidence map into the policy as a task representation (\Cref{sec:algo}). This integration equips the agent with zero-shot generalization ability over objects by grounding the novel object in a comprehensible visual representation.

\subsection{Visual Grounding}
\label{sec:seg}

\begin{figure}[t]
    \centering
    \includegraphics[width=\columnwidth]{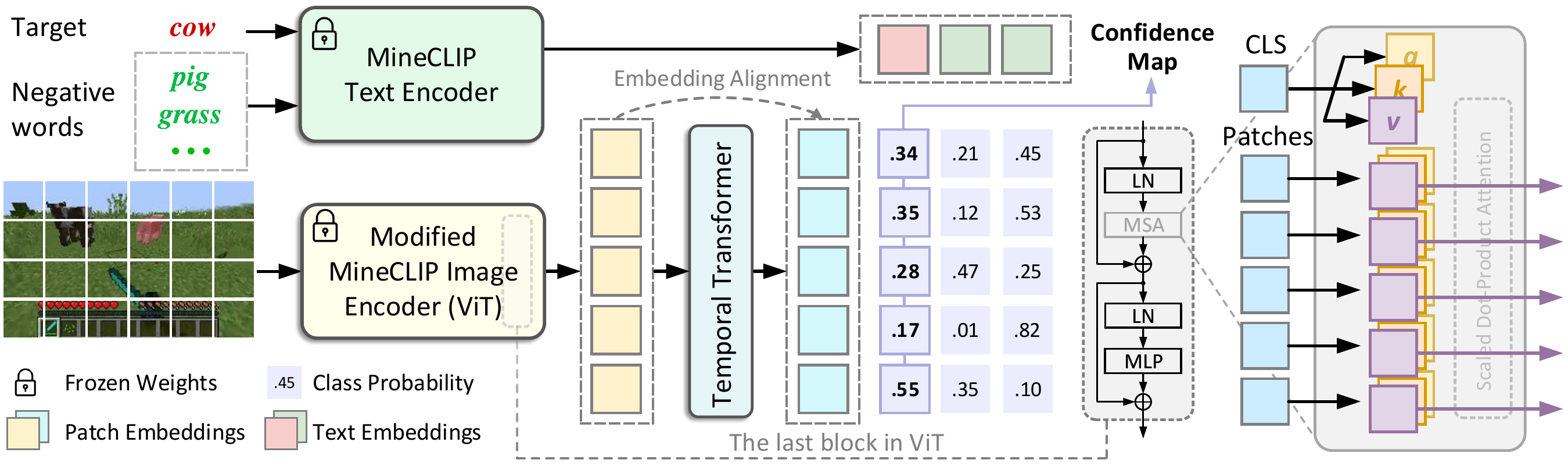}
    \caption{Process of segmentation via MineCLIP. The modified MineCLIP image encoder takes as input the image and outputs patch embeddings, which are subsequently processed by the temporal transformer to guarantee embedding alignment. The MineCLIP text encoder encodes the target name along with a list of negative words. The probability of the target's presence on each patch is calculated based on the similarities between patch embeddings and text embeddings.}
    \label{fig:maskclip}
    \vspace{-2mm}
\end{figure}

Prior to segmentation, we must extract the correct target that the agent needs to interact with from the provided language instruction. Consider an example instruction: ``\texttt{hunt a cow in plains with a diamond sword}''. In this case, it is \texttt{cow} that should be extracted from the instruction as the target object, rather than \texttt{plains} or \texttt{diamond sword}, for the following segmentation. This can be easily accomplished by LLMs with appropriate prompts. Details on prompt designing and conversations with GPT-4 \cite{achiam2023gpt} can be found in \Cref{app:chatgpt}.

In the standard CLIP \cite{radford2021learning}, the image encoder, a ResNet \cite{he2016deep} or ViT \cite{dosovitskiy2020image}, aggregates the visual features from all spatial locations through attention pooling. Recent works \cite{zhou2022extract,li2023clip} reveal that these features on each spatial location contain rich local information so that they can be used to perform zero-shot pixel-level predictions. In brief, the cosine similarities between these features and the outputs of the CLIP text encoder are also valid and informative. Concretely, MaskCLIP \cite{zhou2022extract} makes use of the value-embedding of each spatial location in the last attention module, while CLIPSurgery \cite{li2023clip} studies the feature of each spatial location in the final output and introduces an additional path. Inspired by MaskCLIP, we make adaptations to MineCLIP architecture to generate a confidence map for a specified target \emph{without fine-tuning}.

To begin, we introduce the modification to the vision pathway of MineCLIP. We make changes to extract dense features from the last block of ViT. As illustrated in the rightmost part of \Cref{fig:maskclip}, the scaled dot-product attention in multi-head attention \cite{vaswani2017attention} module is removed, while the \emph{value-embedding transformation} is retained. Then the transformed embeddings excluding that of \texttt{CLS} token are fed into the remaining modules within the ViT to obtain the final embedding of each patch. In this way, these patch embeddings share the same space as the original ViT output. As shown in \Cref{fig:maskclip}, the modified image encoder outputs patch embeddings instead of image embedding. However, these embeddings are not yet aligned with the embedding space of MineCLIP. In MineCLIP, the image encoder is followed by a temporal transformer that aggregates the embeddings of 16 images. Therefore, these patch embeddings also need to pass through the temporal transformer to guarantee alignment. Notably, these embeddings do not form a temporal sequence together as the input of the transformer. Instead, each patch embedding is individually processed by the temporal transformer, treated as a sequence of length 1. In this way, we obtain patch embeddings in the MineCLIP embedding space.

\begin{figure}[t]
    \centering
    \begin{subfigure}{0.3\textwidth}
        \includegraphics[width=1.0\textwidth]{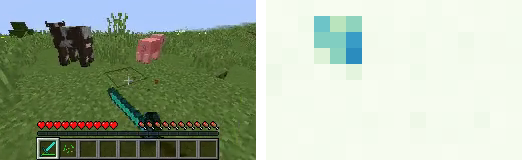}
        \caption{\texttt{cow}}
        \vspace{0.1in}
    \end{subfigure}
    \quad
    \begin{subfigure}{0.3\textwidth}
        \includegraphics[width=1.0\textwidth]{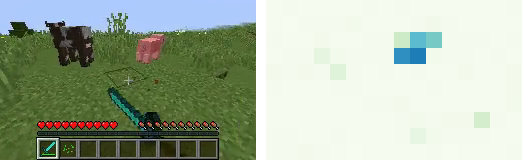}
        \caption{\texttt{pig}}
        \vspace{0.1in}
    \end{subfigure}
    \quad
    \begin{subfigure}{0.3\textwidth}
        \includegraphics[width=1.0\textwidth]{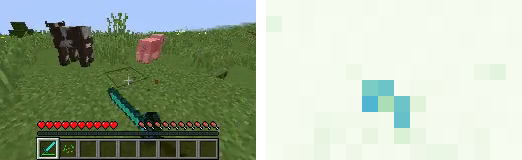}
        \caption{\texttt{sword}}
        \vspace{0.1in}
    \end{subfigure}
    \begin{subfigure}{0.3\textwidth}
        \includegraphics[width=1.0\textwidth]{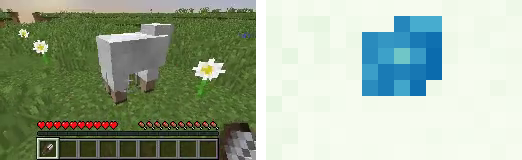}
        \caption{\texttt{sheep}}
    \end{subfigure}
    \quad
    \begin{subfigure}{0.3\textwidth}
        \includegraphics[width=1.0\textwidth]{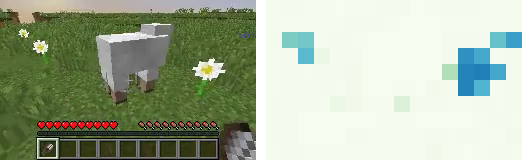}
        \caption{\texttt{flower}}
    \end{subfigure}
    \quad
    \begin{subfigure}{0.3\textwidth}
        \includegraphics[width=1.0\textwidth]{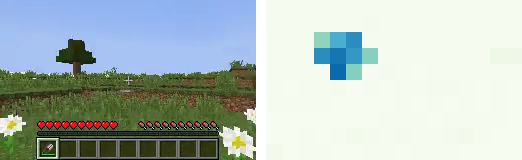}
        \caption{\texttt{tree}}
    \end{subfigure}
    \caption{Segmentation instances for targets: (a) \texttt{cow}, (b) \texttt{pig}, (c) \texttt{sword}, (d) \texttt{sheep}, (e) \texttt{flower}, and (f) \texttt{tree}. 
    The darker blue the patch, the higher the probability of the target's presence on it.}
    \label{fig:seg_eg}
    \vspace{-2mm}
\end{figure}

In the language pathway, no modification is made to the MineCLIP text encoder. The target name is encoded using the text encoder, along with a list of negative words \cite{li2022grounded,zhou2022extract}. We construct a negative word list containing objects that frequently appear in Minecraft. For a detailed description of the word list, please refer to \Cref{app:neg}. Given the patch embeddings encoded through the modified image encoder and the temporal transformer in the same embedding space of MineCLIP, we can calculate cosine similarities between patch embeddings and text embeddings, following the same approach as CLIP. Subsequently, we use softmax with the same temperature used in MineCLIP to determine the probabilities of objects' presence on each patch. Finally, we extract and reshape the probabilities of the target object to form the confidence map. The resulting confidence map consists of the same number of elements as the patches, with each element representing the probability of the target's presence on the corresponding patch. Examples of the confidence maps are shown in \Cref{fig:seg_eg}. 

In our preliminary experiment, we qualitatively attempt off-the-shelf open-vocabulary detection models \cite{kirillov2023segany,liu2023grounding} but find their performance to be impaired by the domain gap between Minecraft and the real world, not as satisfactory as the results of our domain-specific model modified based on MineCLIP, as demonstrated in \Cref{app:dino}. Another advantage of our method is that the generation of the confidence map can be integrated with the calculation of MineCLIP reward or the encoding of images using MineCLIP encoder, thus avoiding significant computational costs from incorporating additional segmentation models.

\subsection{Transfer via Reward}
\label{sec:reward}

\begin{figure}[t]
    \centering
    \begin{subfigure}{0.3\textwidth}
        \includegraphics[width=1.0\textwidth]{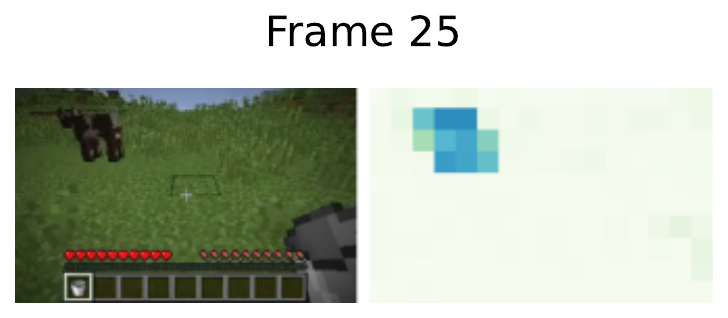}
        \caption{$r^{mc}: 0.856$, $r^{f}:0.029$}
    \end{subfigure}
    \quad
    \begin{subfigure}{0.3\textwidth}
        \includegraphics[width=1.0\columnwidth]{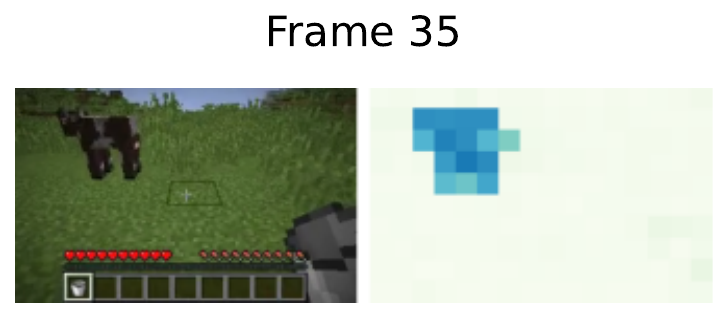}
        \caption{$r^{mc}: 0.574$, $r^{f}:0.046$}
    \end{subfigure}
    \quad
    \begin{subfigure}{0.3\textwidth}
        \includegraphics[width=1.0\columnwidth]{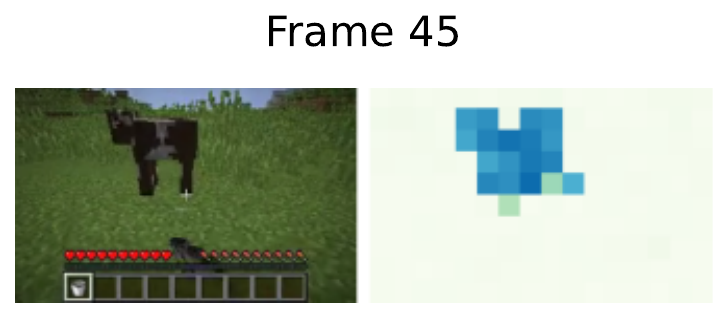}
        \caption{$r^{mc}: 0.600$, $r^{f}:0.071$}
    \end{subfigure}
    \caption{Comparison between MineCLIP reward $r^{mc}$ and focal reward $r^{f}$ at Frame 25, 35, and 45, in one episode of the task \texttt{milk a cow}. From (a) to (c), our focal reward consistently increases as the agent approaches the target \texttt{cow}, while the MineCLIP reward varies in an uncorrelated way.}
    \vspace{-2mm}
    \label{fig:rew_cmp}
\end{figure}

The object-grounded confidence map of the target contains rich spatial information that can be utilized to facilitate RL through reward designing. The area occupied by the target in the image can serve as a \emph{proxy} for estimating the distance to the target, based on the principle that the closer the target is to the agent, the larger its area in the image and vice versa. Therefore, a reward proportional to the area of the target would guide the agent towards the target effectively. Additionally, we argue that the agent should be encouraged to aim at the target, \ie, adjust the perspective to center the target in the field of view. This would help the agent further stabilize its orientation and increase the chance of interacting with the target when it is close enough. In Minecraft, interaction can only occur when the crosshair in the center of the agent view aligns with the target. Moreover, when multiple target objects are present in the view, the agent should learn to focus on a single target rather than attempting to keep all of them in view. This could also be interpreted in a more general way, such as humans usually place the target at the center of the visual field for better perception and interaction.

Based on these principles, we introduce an object-grounded intrinsic reward function named \emph{focal} reward. At each time step $t$, it is computed as the mean of the Hadamard product between the current target confidence map $m^c_t$, and a Gaussian kernel denoted as $m^k$:
\begin{equation}
    \label{equ:focal}
    r^{f}_t = \operatorname{mean}\left(m^c_t\circ m^k\right).
\end{equation}
Here, $m^c_t$ and $m^k$ share the same dimensions with height $H$ and width $W$. Each element of the Gaussian kernel is defined as: 
\begin{equation}
    m^k_{i,j} = \exp\left({-\frac{(i-\mu_1)^2}{2\sigma^2_1}}-\frac{(j-\mu_2)^2}{2\sigma^2_2}\right),
\end{equation}
where $\mu_1 = (H+1)/2$, $\sigma_1 = H/3$, $\mu_2 = (W+1)/2$, and $\sigma_2 = W/3$. This reward function is designed to be directly proportional to the area occupied by the target and inversely proportional to the distance between the target patches and the center of the view. 

As noted in \cite{cai2023open}, the MineCLIP reward, which relies on the similarity between the agent's preceding image observations and the provided instruction, is uncorrelated with the distance between the agent and the target. This phenomenon is demonstrated in \Cref{fig:rew_cmp}, where the MineCLIP reward does not consistently increase as the agent gets closer to the target. Consequently, in practice, the agent trained with the MineCLIP reward tends to ``stare at'' the target at a distance, rather than approaching it. This tendency obstructs the agent from learning some hard-exploration skills, particularly those that require multiple times of interactions with the targets, such as hunting. In contrast, our focal reward addresses this deficiency, increasing consistently when the agent approaches the target \texttt{cow}, as illustrated in \Cref{fig:rew_cmp}.

The confidence map generated from the modified MineCLIP may sometimes contain noisy activation \cite{zhou2022extract,li2023clip}. Therefore, we process the raw confidence map to enhance its quality before using it to compute the intrinsic reward. Firstly, we set the value corresponding to the patch where a word from the negative word list has the highest probability instead of the target, to zero. This operation reduces the impact of noisy activation on non-target patches. Secondly, we set values in the confidence map lower than a threshold $\tau=0.2$ to zero, while those higher than this threshold are set to one, so as to amplify the distinction between patches corresponding to the target and those unrelated to it. We ablate the Gaussian kernel and denoising process in Section \ref{sec:single}.

\subsection{Transfer via Representation}
\label{sec:algo}

To train an instruction-following agent, the conventional practice involves directly taking the natural language instruction as the task representation into the policy network \cite{jiang2019language,khandelwal2022simple,mu2022improving,du2023guiding}. These instructions are typically encoded using a recurrent network or a language model such as BERT \cite{kenton2019bert} and CLIP \cite{radford2021learning}. In contrast, we extract the target object from the instruction using GPT-4 \cite{achiam2023gpt} and subsequently convert it into a two-dimensional matrix, \ie, the confidence map. Our underlying assumption is that this two-dimensional object-grounded representation offers more intuitive and accessible information for the policy network compared to the intricate space of language embeddings. When facing an instruction containing the name of an target object not encountered during training, our method grounds this novel text in the two-dimensional map, rendering it a comprehensible representation for the policy network. As a result, the agent can follow the guidance of the confidence map, navigate towards the novel target object, and finally interact with it. 

In our implementation, we adopt the network architecture of MineAgent \cite{fan2022minedojo}, which uses the MineCLIP image encoder to process image observations and MLPs to encode other information such as pose. We introduce an additional branch to encode the confidence map and fuse these features through concatenation. The policy network takes this fused multi-modality feature as input and outputs action distribution. Details regarding the policy network's architecture are available in \Cref{app:net}. We use PPO \cite{schulman2017proximal} as the base RL algorithm and train the agent with reward $r_t=r^{env}_t+\lambda r^{f}_t$, where $r^{env}$ denotes the environmental reward and $\lambda$ is a hyperparameter controlling the weight of the focal reward. According to the empirical results in \Cref{app:coef}, we simply set $\lambda=5$ for all experiments in the paper as we do not want to bother tuning this hyperparameter. We employ the multi-task RL paradigm, where the agent is trained to finish tasks in a predefined instruction set. Unlike typical multi-task reinforcement learning, our agent's learning objective is to not only master the training tasks but also to understand the mapping between the confidence map and the target object within the image observation, in order to perform zero-shot generalization to novel instructions involving unseen target objects.

\section{Experiments}

We conduct experiments in MineDojo \cite{fan2022minedojo}, a Minecraft simulator that offers diverse open-ended tasks. Firstly, we perform single-task experiments to evaluate the effectiveness of our proposed focal reward. Then we extend our evaluation to multi-task experiments and examine the performance of COPL on multiple instructions. Lastly, but most importantly, we investigate the zero-shot generalization ability of COPL when confronted with instructions containing unseen targets. Details about Minecraft environments and RL hyperparameters in our experiments are described in \Cref{app:space} and \Cref{app:ppo}, respectively.

\subsection{Single-Task Experiments}
\label{sec:single}

Our single-task evaluation consists of tasks learning four challenging basic skills\cite{baker2022video,fan2022minedojo,cai2023open}: \texttt{hunt a cow}, \texttt{hunt a sheep}, \texttt{hunt a pig}, and \texttt{hunt a chicken}. In each task, the agent spawns in \texttt{plains} biome alongside several animals. The agent will receive a reward from the environment if it successfully kills the target animal. The difficulty of these basic skills lies in that animals, once attacked, will attempt to flee, requiring the agent to keep chasing and attacking the target animal. More details about the Minecraft task settings are available in \Cref{app:env_single}. 

\vspace{1mm}
\noindent \textbf{Evaluation.} We compare our focal reward with the following three baselines: (1) \textbf{MineCLIP reward} \cite{fan2022minedojo} based on the similarity score between image observations and the instruction ``\texttt{hunt a \{animal\} on plains with a diamond sword}''; (2) \textbf{ND\textsubscript{CLIP} reward} \cite{tam2022semantic}, an intrinsic reward for exploration that measures the novelty of observation's MineCLIP embedding; (3) \textbf{Sparse reward}, training the agent with the environmental reward only. Results are reported in \Cref{tab:single}, including mean and variance, calculated from evaluating four models that are each trained with a unique random seed and the same number of environment steps (the same applies hereinafter). We can observe that only our focal reward leads to the mastery of all four skills by guiding the agent to consistently approach the target. In contrast, the MineCLIP reward fails because it cannot capture the distance between the agent and the target, offering limited benefit to these tasks. The failure of ND\textsubscript{CLIP} reward suggests that exploration provides minimal assistance in learning these challenging skills due to the huge observation space of Minecraft. These methods' learning curves on each task are available in \Cref{app:single_curve}. We also report results on harvest skill learning and additional analysis in \Cref{app:single_more}.

\vspace{1mm}
\noindent \textbf{Variants and Ablation.} To further investigate our focal reward, we compare it with two variants: Focal[raw], which uses the raw confidence map without denoising to compute the intrinsic reward, and Focal[delta], defined as $r^\delta_t=r^{f}_t-r^{f}_{t-1}$. The results in \Cref{tab:abltn} demonstrate that our denoising process improves the effectiveness of the focal reward. We suppose that the poor performance of Focal[delta] may be linked to its sensitivity to segmentation noise, as it relies on differences in focal reward between two steps, making it susceptible to minor fluctuations in segmentation. In addition, we test the effectiveness of the Gaussian kernel, as presented in \Cref{tab:abltn}. We modify the environment settings to ensure that there are two target animals. The results prove the significance of the Gaussian kernel. Without this kernel, the reward may guide the agent to include both target animals in the view to acquire a high reward, hindering it from approaching either of them. In contrast, our focal reward addresses this problem by providing more reward in the center, thereby encouraging the agent to focus on a single target. A parameter study for the variance of the Gaussian kernel is available in \Cref{app:gauss}.

\begin{table}[t]
\centering
\renewcommand{\arraystretch}{1.2}
\caption{Success rates (\%) of single-task RL with different reward functions on four challenging Minecraft tasks. Our focal reward enables RL to master all tasks by guiding the agent to consistently approach the target, while other baselines fail.}
\begin{footnotesize}
\begin{tabular}{p{3pt}lp{5pt}lp{3pt}lp{3pt}lp{3pt}lp{3pt}}
\toprule
& \multicolumn{1}{c}{\textbf{Tasks}} & & \multicolumn{1}{c}{Focal} & & \multicolumn{1}{c}{MineCLIP} & & \multicolumn{1}{c}{ND\textsubscript{CLIP}} & & \multicolumn{1}{c}{Sparse} & \\ \midrule
& \texttt{hunt a cow}               & & \textbf{71.3±9.7}  & & 3.8±4.8 & & 3.5±3.0 & & 0.0±0.0 & \\
& \texttt{hunt a sheep}             & & \textbf{68.8±25.3} & & 5.3±2.9 & & 28.8±23.0 & & 2.5±3.0 & \\
& \texttt{hunt a pig}               & & \textbf{58.3±7.8}  & & 2.3±1.7 & & 0.3±0.5 & & 0.5±0.6 & \\
& \texttt{hunt a chicken}           & & \textbf{29.5±10.9} & & 0.0±0.0 & & 4.8±1.5 & & 0.5±0.6 & \\ \bottomrule
\end{tabular}
\end{footnotesize}
\label{tab:single}
\end{table}

\begin{table}[t]
\caption{Success rates (\%) of single-task RL with variants and ablation of our focal reward on two Minecraft tasks.}
\centering
\renewcommand{\arraystretch}{1.2}
\begin{tabular}{p{3pt}lp{5pt}lp{3pt}lp{3pt}lp{3pt}lp{3pt}lp{3pt}}
\toprule
& \multicolumn{1}{c}{\multirow{2}{*}{\textbf{Tasks}}} & & \multicolumn{5}{c}{\textbf{Variants}} & & \multicolumn{3}{c}{\textbf{Ablation}} & \\ \cline{4-8} \cline{10-12}
\rule{0pt}{2.8ex} & & & \multicolumn{1}{c}{Focal} & & \multicolumn{1}{c}{[raw]} & & \multicolumn{1}{c}{[delta]} & & \multicolumn{1}{c}{Focal} & & \multicolumn{1}{c}{w/o kernel} & \\ \midrule
& \texttt{hunt a cow} & & \textbf{71.3±9.7} & & 3.8±4.8 & & 3.5±3.0 & & \textbf{67.0±15.0} & & 14.5±21.8 & \\
& \texttt{hunt a sheep} & & \textbf{68.8±25.3} & & 5.3±2.9 & & 28.8±23.0 & & \textbf{82.3±3.4} & & 59.8±23.0 & \\ \bottomrule
\end{tabular}
\label{tab:abltn}
\end{table}

\subsection{Multi-Task and Generalization Experiments}
\label{sec:multi}

We conduct multi-task experiments to verify the effectiveness and zero-shot generalization capability of COPL. Given that tasks in Minecraft require different behavior patterns, we design two task domains, the \textbf{hunting domain} and the \textbf{harvest domain}. The hunting domain consists of four instructions: ``\texttt{hunt a cow}'', ``\texttt{hunt a sheep}'', ``\texttt{hunt a pig}'', and ``\texttt{hunt a chicken}''. These tasks share a common behavior pattern: repeatedly \textit{approach the target}, \textit{aim at it}, and \textit{attack}. The harvest domain also contains four instructions: ``\texttt{milk a cow}'', ``\texttt{shear a sheep}'', ``\texttt{harvest a flower}'', and ``\texttt{harvest leaves}''. Tasks in the harvest domain are individually easier than those in the hunting domain but demand disparate behavior patterns. For example, ``\texttt{harvest a flower}'' requires the \textit{attack} action while the other tasks require the \textit{use} action. More details about the task settings in multi-task experiments are available in \Cref{app:env_multi}.

\begin{table}[t]
    \centering
    \renewcommand{\arraystretch}{1.2}
    \caption{Success rates (\%) of different multi-task RL methods in the hunting domain. (\textit{upper}) Training hunting tasks. (\textit{lower}) Test hunting tasks on \textbf{unseen} targets. By visual grounding, COPL achieves 4$\times$ the performance of language-conditioned methods on unseen targets, demonstrating zero-shot object-level generalization ability.}
    \begin{tabular}{p{1pt}lp{3pt}lp{3pt}lp{1pt}lp{1pt}lp{3pt}lp{3pt}lp{1pt}} \toprule
     & \multicolumn{1}{c}{\textbf{Tasks}} & & \multicolumn{1}{c}{COPL} &  & \multicolumn{1}{c}{LCRL[t]} & & \multicolumn{1}{c}{LCRL[i]} & & \multicolumn{1}{c}{One-Hot\footnotemark} & & \multicolumn{1}{c}{\cite{cai2023open}} & & \multicolumn{1}{c}{STEVE-1} & \\ \midrule
     & \texttt{cow}     & & \textbf{55.5±13.6}  & & 18.8±22.2   & & 12.8±8.5    & & 23.0±20.3   & & 31.0    & & 6.0     & \\
     & \texttt{sheep}   & & \textbf{60.0±3.9}   & & 31.8±25.2   & & 24.3±17.7   & & 31.0±12.7   & & 30.0    & & 14.0    & \\
     & \texttt{pig}     & & \textbf{50.5±18.4}  & & 8.8±5.8     & & 17.0±15.9   & & 3.3±1.0     & & 34.0    & & 9.0     & \\
     & \texttt{chicken} & & \textbf{34.0±3.2}   & & 7.3±8.8     & & 10.3±4.9    & & 11.5±7.9    & & 1.0     & & 6.0     & \\
     & Avg.             & & \textbf{50.0±8.9}   & & 16.6±13.3   & & 16.1±8.8    & & 17.2±7.9    & & 24.0    & & 8.8     & \\ \midrule
     & \texttt{llama}   & & \textbf{48.8±6.5}   & & 14.5±10.4   & & 24.5±12.7   & & \multicolumn{1}{c}{-} & & 11.0    & & 2.0 & \\
     & \texttt{horse}   & & \textbf{49.0±5.5}   & & 2.5±1.3     & & 5.5±4.7     & & \multicolumn{1}{c}{-} & & 7.0     & & 3.0 & \\
     & \texttt{spider}  & & 54.5±12.7   & & 9.8±3.5     & & 18.3±12.0   & & \multicolumn{1}{c}{-} & & 7.0 & & \textbf{73.0}   & \\
     & \texttt{mushroom} & & \textbf{40.3±11.2} & & 19.3±20.5   & & 0.0±0.0     & & \multicolumn{1}{c}{-} & & 26.0    & & 4.0 & \\
     & Avg.             & & \textbf{48.1±6.6}   & & 11.5±7.6    & & 12.1±4.9    & & \multicolumn{1}{c}{-} & & 12.8   & & 20.5    & \\ \bottomrule
    \end{tabular}
\label{tab:mt_combat}
\end{table}
\footnotetext{One-Hot is not evaluated on unseen tasks due to the absence of corresponding one-hot vectors outside the training set.}

\vspace{1mm}
\noindent \textbf{Evaluation.} We compare COPL with two language-conditioned reinforcement learning (LCRL) baselines, both taking language as input \cite{jiang2019language,luketina2019survey,khandelwal2022simple,mu2022improving}: (1) \textbf{LCRL[t]}, which utilizes the target embedding encoded by the MineCLIP text encoder as input, sharing the same information as COPL albeit in language; (2) \textbf{LCRL[i]}, which utilizes the instruction embedding encoded by the MineCLIP text encoder as input. We also evaluate (3) \textbf{One-Hot}, a naive multi-task baseline, using a one-hot vector as the task indicator. All these methods are trained with the focal reward and the only difference is their conditioning task representations. In the hunting domain, as shown in \Cref{tab:mt_combat} (\textit{upper}), COPL significantly outperforms other baselines, indicating that the confidence map provides a more accessible and informative task representation compared to the language embedding and one-hot vector, respectively. In contrast, the harvest domain presents a different picture. As illustrated in \Cref{tab:mt_harvest} (\textit{upper}), all methods achieve similar performance. These results suggest that when tasks become easy enough, the impact of the task representation's complexity diminishes. These methods' learning curves on each task are available in \Cref{app:mt_rl}. We also evaluate two recent Minecraft basic skill models trained via imitation learning, \cite{cai2023open}\footnote{\cite{cai2023open} is not evaluated in the harvest domain because the authors have not yet released the model trained for harvest tasks.} and STEVE-1\cite{lifshitz2023steve}. Note that our intention \emph{is not} to directly compare the performance of COPL with that of these two models, given the significant differences in their training paradigms and scopes of data. Rather, their evaluations only serve as a reference here due to the lack of alternative Minecraft foundation models trained via RL.

\vspace{1mm}
\noindent \textbf{Generalization.} Given that the two domains involve distinct behavior patterns, we conduct separate evaluations to assess the zero-shot generalization ability over target objects of these models trained in the hunting domain and the harvest domain. We test the hunting models with four novel instructions involving unseen animals: ``\texttt{hunt a llama}'', ``\texttt{hunt a horse}'', ``\texttt{hunt a spider}'', and ``\texttt{hunt a mushroom cow}''. The results in \Cref{tab:mt_combat} (\textit{lower}) show that COPL effectively transfers the learned skill to unseen targets, achieving high success rates. As we set the episode to be terminated if any animal is killed, the high success rates also prove COPL's ability to distinguish the target from other animals, rather than indiscriminately attacking them. Detailed precision of each method is reported in \Cref{app:prcsn}. STEVE-1 shows poor performance across all hunting tasks except ``\texttt{hunt a spider}''. We suppose that its base model, VPT \cite{baker2022video}, possesses a strong prior on killing specific animals like spiders and heavily affects the behavior of STEVE-1 on following other hunting instructions. \cite{cai2023open} achieves relatively higher success rates on ``\texttt{hunt a cow}'', ``\texttt{hunt a sheep}'', and ``\texttt{hunt a pig}'' due to these tasks being in its training set. Its lower performance on other tasks indicates its limitations in generalization ability. 

\begin{table}[t]
    \centering
    \renewcommand{\arraystretch}{1.2}
    \caption{Success rates (\%) of different multi-task RL methods in the harvest domain. (\textit{upper}) Training harvest tasks. (\textit{lower}) Test harvest tasks on \textbf{unseen} targets. By visual grounding, COPL achieves 2$\times$ the performance of language-conditioned methods on unseen targets, demonstrating zero-shot object-level generalization ability.}
    \begin{tabular}{p{2pt}lp{3pt}lp{3pt}lp{3pt}lp{3pt}lp{3pt}lp{2pt}} \toprule
     & \multicolumn{1}{c}{\textbf{Tasks}} & & \multicolumn{1}{c}{COPL} &  & \multicolumn{1}{c}{LCRL[t]} & & \multicolumn{1}{c}{LCRL[i]} & & \multicolumn{1}{c}{One-Hot} & & \multicolumn{1}{c}{STEVE-1} & \\ \midrule
     & \texttt{milk}    & & \textbf{77.0±3.9}  & & 68.0±4.6   & & 73.5±11.7  & & 69.5±5.0   & & 2.0   & \\
     & \texttt{wool}    & & \textbf{55.5±5.7}  & & 54.5±13.0  & & 53.0±6.4   & & 55.0±9.3   & & 7.0   & \\
     & \texttt{flower}  & & 85.5±7.7 & & \textbf{86.5±5.3}    & & 81.5±6.5   & & 80.3±6.8   & & 43.0  & \\
     & \texttt{leaves}  & & \textbf{84.5±4.0}  & & 83.7±2.5   & & 73.7±5.9   & & 74.0±5.9   & & 66.0  & \\
     & Avg.             & & \textbf{75.6±3.2}  & & 73.2±5.2   & & 70.4±2.0   & & 69.7±2.5   & & 29.5  & \\ \midrule
     & \texttt{water}   & & \textbf{46.5±12.7} & & 25.8±10.3  & & 25.8±3.3   & & \multicolumn{1}{c}{-} & & 21.0 & \\
     & \texttt{mushroom}& & \textbf{38.8±7.5}  & & 29.0±6.2   & & 31.3±3.1   & & \multicolumn{1}{c}{-} & & 0.0  & \\
     & \texttt{sand}    & & \textbf{24.0±8.4} & & 2.5±2.9    & & 3.5±4.5    & & \multicolumn{1}{c}{-} & & 11.0 & \\
     & \texttt{dirt}    & & 54.0±24.2   & & 9.5±9.5 & & 18.3±21.2   & & \multicolumn{1}{c}{-}   & & \textbf{75.0}   & \\
     & Avg.             & & \textbf{40.8±8.3}  & & 16.7±4.3   & & 19.7±6.3   & & \multicolumn{1}{c}{-} & & 26.8 & \\ \bottomrule
    \end{tabular}
\label{tab:mt_harvest}
\end{table}

For the harvest domain, we conduct evaluation on four unseen instructions, including ``\texttt{harvest water}'', ``\texttt{shear a mushroom cow}'', ``\texttt{collect sand}'', and \\``\texttt{collect dirt}''. As shown in \Cref{tab:mt_harvest} (\textit{lower}), COPL exhibits advantages on unseen tasks, reaching higher success rates than LCRL[t] and LCRL[i]. Compared to the hunting domain, results here more clearly demonstrate our method's superior generalization ability over target objects, as all methods perform at the same level on the training tasks. This indicates that better generalization ability emerges from grounding language in a visual representation. Evidenced by the precision reported in \Cref{app:prcsn}, COPL recognizes unseen targets more accurately compared to LCRL. Similar to our observation in the hunting domain, STEVE-1 also possesses a strong tendency, which is digging in this case.

\section{Discussion}

\textbf{Beyond Object-Centric Tasks.} We focus on object-centric tasks in this paper, employing an intrinsic reward function to guide a learning agent towards the target. However, not all tasks in Minecraft necessitate approaching the target, such as the two Creative tasks, \texttt{dig a hole} and \texttt{lay the carpet}, adopted in \cite{fan2022minedojo}. We explore training separate single-task policies using the focal reward with targets, 
\begin{wrapfigure}{r}{0.62\textwidth}
    \centering
    \begin{subfigure}{0.3\textwidth}
        \includegraphics[width=1.0\textwidth]{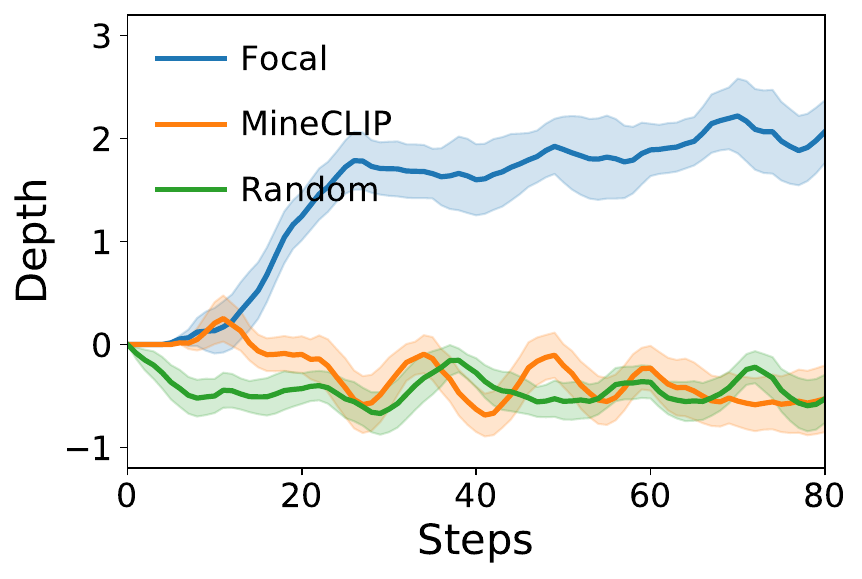}
        \caption{\texttt{dig a hole}}
        \label{fig:dig}
    \end{subfigure}
    \begin{subfigure}{0.3\textwidth}
        \includegraphics[width=1.0\textwidth]{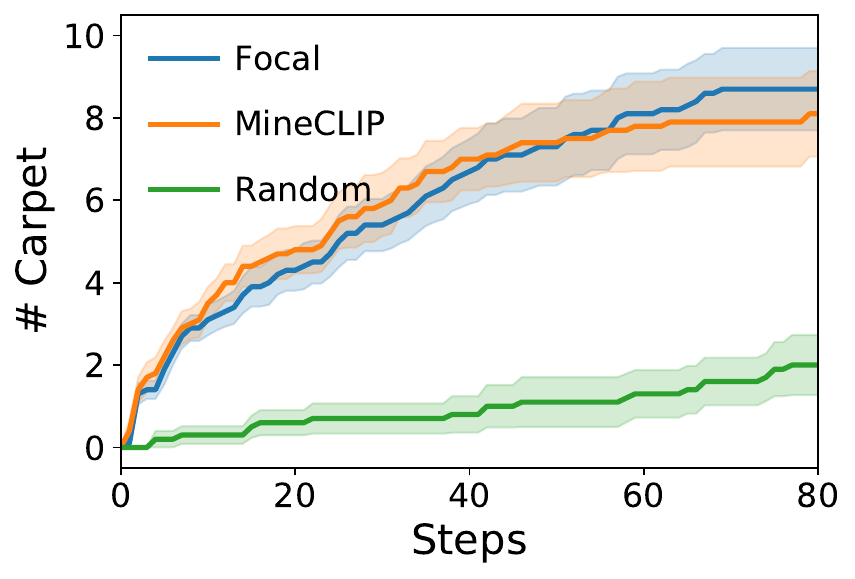}
        \caption{\texttt{lay the carpet}}
        \label{fig:carpet}
    \end{subfigure}
    \caption{(a) Digging depth and (b) number of laid carpets in one episode. The increase in the y-axis metric with steps indicates that our focal reward successfully guides the agent to conduct the corresponding tasks.}
    \label{fig:creative}
\end{wrapfigure}
\texttt{hole} and \texttt{carpet}, respectively. Results in \Cref{fig:creative} show effective guidance towards task completion. Nevertheless, for zero-shot generalization, we observe that both our hunting and harvest models show limited performance on \texttt{dig a hole} due to its different behavior pattern from the training tasks and the difficulty in selecting a consistent target dirt block. Details on results of the two tasks are further discussed in \Cref{app:creative}.

\vspace{1mm}
\noindent \textbf{Limitations.} One limitation of our method is the difficulty in defining a target for COPL to condition on, \ie, for the agent to aim at, when facing non-object-centric tasks, such as \texttt{dig a hole} and \texttt{build a house}. Solving such creative and building tasks may entail generative models\cite{zhang2023creative}. Additionally, our method compromises the \textit{theoretical} generality of MineCLIP \cite{fan2022minedojo} due to not considering the action in instructions and only grounding target objects. Consequently, our method significantly improves its \textit{practicality} in object-centric tasks, as verified in our experiments. Future work could focus on grounding language that describes actions and learning tasks requiring more complicated manipulation.

\section{Conclusion}

We propose COPL, a novel approach designed to address object-centric tasks and perform zero-shot object-level generalization in Minecraft. Our method effectively transfers the wealth of vision-language knowledge encoded in MineCLIP \cite{fan2022minedojo} into RL via reward designing and task representation. By comprehensive evaluations, we prove COPL's effectiveness in acquiring multiple basic skills and its generalization ability over target objects indicated in instructions, enabling it to apply the learned skills to follow unseen instructions that involve objects beyond the training tasks. Our work demonstrates the potential of integrating multimodal models, such as VLMs, into RL. Our method can be applied to other similar open-world environments by grounding natural language instructions in visual data and guiding the agent toward targets likewise. We hope COPL could contribute to the development of agents capable of understanding and responding to natural language instructions.

\section*{Acknowledgements}
This work was supported by NSFC under grant 62250068. The authors would like to thank Ziluo Ding for providing help with the writing of this paper. The authors also would like to thank the anonymous reviewers for their valuable comments. 

%
%
\bibliographystyle{splncs04}
\bibliography{main}

\newpage
\appendix

\section{Segmentation Details}

\subsection{Extracting Targets via LLM}
\label{app:chatgpt}

In order to extract the target name from a natural language instruction, we resort to GPT-4 \cite{achiam2023gpt}. We input the following prompt:

\newenvironment{noindentenv}
{
    \setlength{\parindent}{0pt}
    \setlength{\parskip}{2mm}
}
{}

\begin{noindentenv}

{\ttfamily
You are a helper agent in Minecraft. I will describe a task, and\\ your job is to identify the specific entity (name of blocks, anim-\\als, or items) that I need to directly interact with to complete\\ this task. This means you should tell me what entity I need to aim\\ at with the crosshair in the center of my screen, assuming I have\\ all the necessary items and tools in my inventory. Please provide\\ the name of the target entity.

Requirements:

1. For tasks with multiple possible solutions or those requiring\\ additional context, provide the most straightforward or common\\ solution.

2. Only output the target entity after "Target: ", DO NOT add other examples, descriptions, or explanations.

For example:

Task: "Obtain 8 obsidian blocks with a diamond pickaxe and a water\\ bucket."

Thinking: "To mine obsidian, the player must aim their crosshair at an obsidian block and use a diamond or netherite pickaxe on it. Ob-\\sidian is created when flowing water touches a lava source block.\\ Therefore, the entity to aim at for this task is obsidian."

Target: "Obsidian"

Task: "Obtain milk with an empty bucket."

Thinking: "To fill a bucket with milk, the player must aim their\\ crosshair at a cow and use the empty bucket on it. Therefore, the\\ entity to aim at for this task is a cow."

Target: "Cow"
}

\end{noindentenv}

\vspace{2mm}
We try some instructions without telling the name of the true target entity:

\begin{noindentenv}

{\ttfamily
\underline{We:}

Task: "Harvest wool"

Thinking: 

\underline{GPT-4:}

To harvest wool in Minecraft, you need to aim your crosshair at a\\ sheep and use shears on it. This will allow you to collect wool\\ without harming the sheep.

Target: "Sheep"

\underline{We:}

Task: "Collect beef"

Thinking: 

\underline{GPT-4:}

To collect beef in Minecraft, you need to aim your crosshair at a\\ cow and use a weapon or your hand to kill it. Upon death, the cow\\ will drop beef.

Target: "Cow"

\underline{We:}

Task: "Get mutton as food"

Thinking: 

\underline{GPT-4:}

To get mutton as food in Minecraft, you need to aim your crosshair\\ at a sheep and use a weapon or your hand to kill it. Upon death,\\ the sheep will drop mutton.

Target: "Sheep"
}

\end{noindentenv}

\vspace{2mm}
These results show that GPT-4 understands the underlying relationship between the target item (\texttt{wool}, \texttt{beef}, \texttt{mutton}) and the target animal (\texttt{sheep}, \texttt{cow}, \texttt{sheep}) based on its prior knowledge about Minecraft. Certainly, GPT-4 also works on instructions containing the target entity directly:

\begin{noindentenv}

{\ttfamily
\underline{We:}

Task: "Harvest a flower"

Thinking: 

\underline{GPT-4:}

To harvest a flower in Minecraft, you need to aim your crosshair at the specific flower you want to collect and either use your hand or any tool to break and collect it.

Target: "Flower"

\underline{We:}

Task: "Hunt a cow"

Thinking: 

\underline{GPT-4:}

To hunt a cow in Minecraft, you need to aim your crosshair at a cow and use a weapon or your hand to kill it.

Target: "Cow"
}
    
\end{noindentenv}

\subsection{Negative Words}
\label{app:neg}

For better segmentation, the negative word list should contain names of objects that frequently appear in Minecraft. To this end, we utilize the TF-IDF algorithm to select top-100 words from the subtitles of YouTube videos \cite{fan2022minedojo}, excluding stop words like ``we'' and ``is'', as well as modal particles such as ``yeah'' and ``uh''. Additionally, we filter out verbs and some irrelevant nouns from the top-100 words to reduce noise. The final negative word list is shown below:

\begin{quote}
\noindent {\ttfamily
diamond, block, village, house, iron, farm, chest, dragon, \\ 
redstone, water, tree, zombie, sword, stone, door, armor, lava, \\
fish, portal, chicken, wood, wall, glass, cave, stair, bed,\\ 
torch, fire, creeper, island, food, slab, book, head, button,\\ 
apple, skeleton, potion, spider, egg, pickaxe, arrow, boat, \\
horse, hopper, box, wool, table, seed, cow, brick, trap, dog,\\
bow, dirt, roof, leaves, sand, window, bucket, coal, hole, pig,\\ 
ice, bone, stick, flower, tower, sheep, grass, sky.
}
\end{quote}

Furthermore, in constructing text embeddings, we employ prompt engineering to improve zero-shot ability on classification \cite{radford2021learning}. Same as MaskCLIP \cite{zhou2022extract}, we utilize 85 prompt templates such as ``\texttt{a photo of many \{\}.}". The mean of these embeddings is set to be the text embedding of the target. During segmentation, if the target object already exists in the list, it will be removed from the list in advance.

\subsection{Segmentation Results}
\label{app:seg}

\begin{figure}[t]
    \centering
    \begin{subfigure}{0.3\textwidth}
        \includegraphics[width=1.0\textwidth]{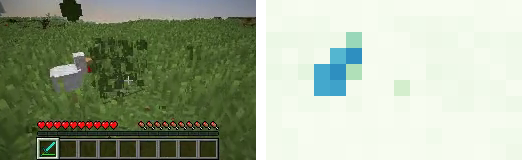}
        \caption{\texttt{chicken}}
        \vspace{0.1in}
    \end{subfigure}
    \begin{subfigure}{0.3\textwidth}
        \includegraphics[width=1.0\textwidth]{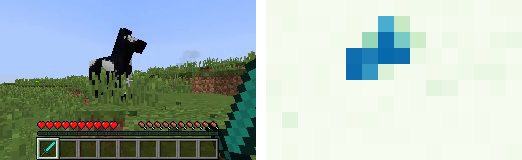}
        \caption{\texttt{horse}}
        \vspace{0.1in}
    \end{subfigure}
    \begin{subfigure}{0.3\textwidth}
        \includegraphics[width=1.0\textwidth]{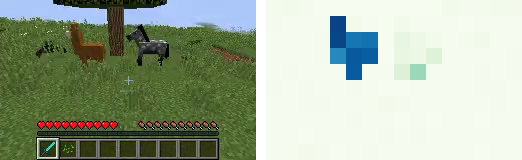}
        \caption{\texttt{llama}}
        \vspace{0.1in}
    \end{subfigure}
    \begin{subfigure}{0.3\textwidth}
        \includegraphics[width=1.0\textwidth]{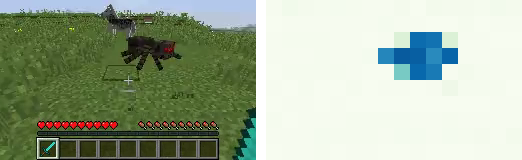}
        \caption{\texttt{spider}}
    \end{subfigure}
    \begin{subfigure}{0.3\textwidth}
        \includegraphics[width=1.0\textwidth]{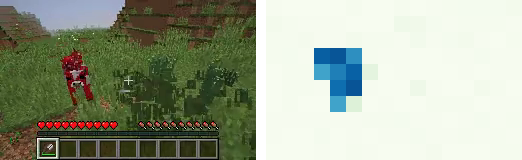}
        \caption{\texttt{mushroom cow}}
    \end{subfigure}
    \begin{subfigure}{0.3\textwidth}
        \includegraphics[width=1.0\textwidth]{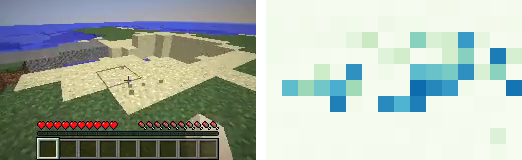}
        \caption{\texttt{sand}}
    \end{subfigure}
    \caption{Confidence map instances of targets: (a) \texttt{chicken}, (b) \texttt{horse}, (c) \texttt{llama}, (d) \texttt{spider}, (e) \texttt{mushroom cow}, and (f) \texttt{sand}.}
    \label{fig:seg_eg_more}
\end{figure}

We provide more examples of confidence maps, as illustrated in \Cref{fig:seg_eg_more}. Our modified MineCLIP effectively locates these target objects.

\subsection{Off-the-shelf Object Detection Models}
\label{app:dino}

\begin{figure}[t]
    \centering
    \includegraphics[width=0.45\columnwidth]{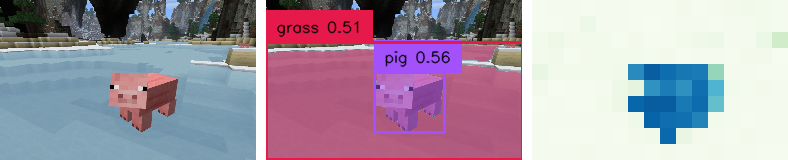}
    \hspace{3mm}
    \includegraphics[width=0.45\columnwidth]{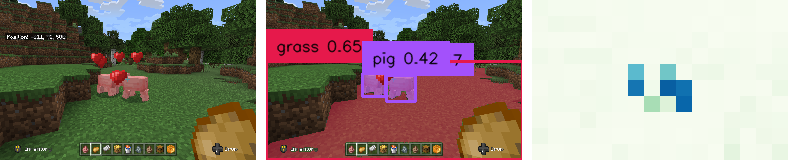}
    \includegraphics[width=0.45\columnwidth]{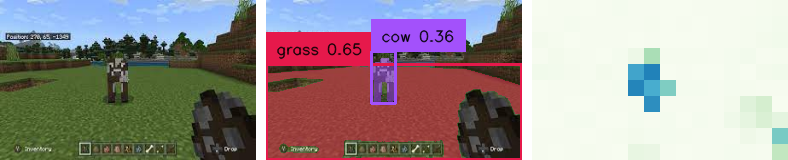}
    \hspace{3mm}
    \includegraphics[width=0.45\columnwidth]{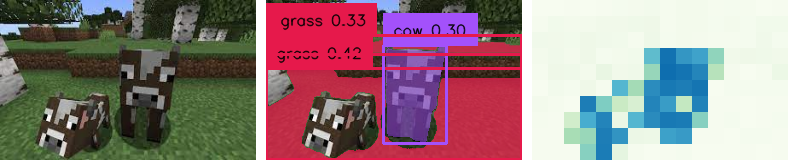}
    \includegraphics[width=0.45\columnwidth]{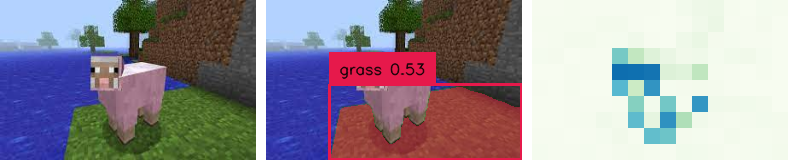}
    \hspace{3mm}
    \includegraphics[width=0.45\columnwidth]{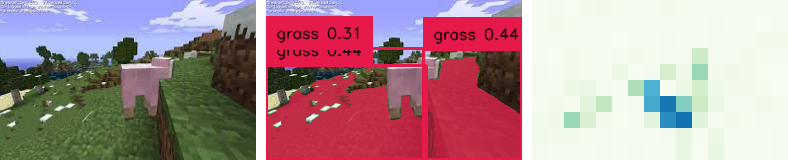}
    \includegraphics[width=0.45\columnwidth]{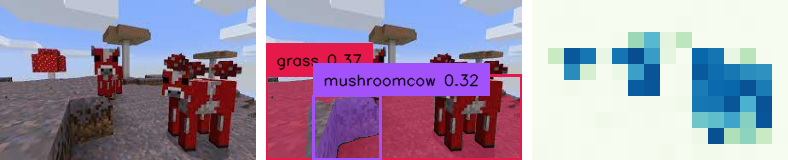}
    \hspace{3mm}
    \includegraphics[width=0.45\columnwidth]{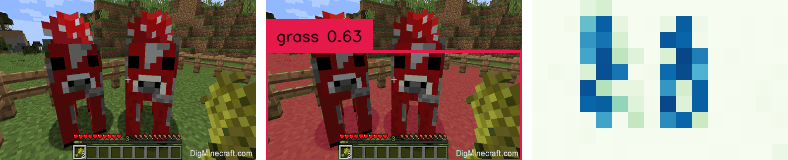}
    \includegraphics[width=0.45\columnwidth]{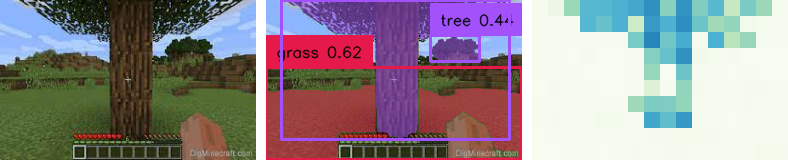}
    \hspace{3mm}
    \includegraphics[width=0.45\columnwidth]{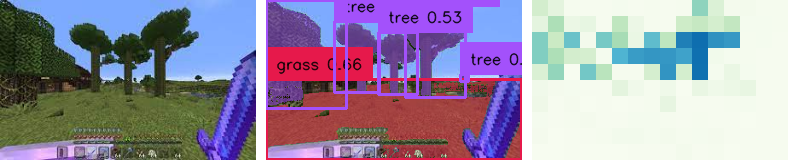}
    \includegraphics[width=0.45\columnwidth]{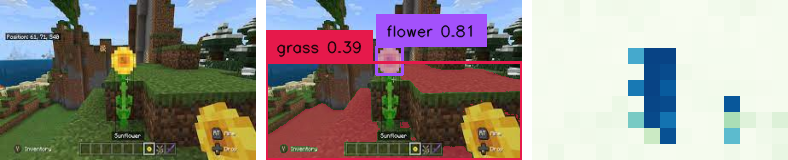}
    \hspace{3mm}
    \includegraphics[width=0.45\columnwidth]{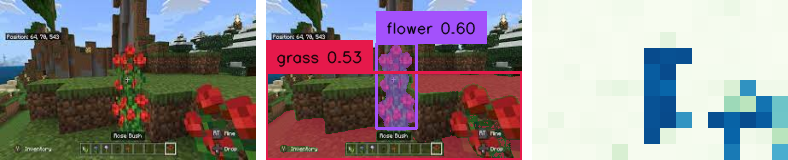}
    \includegraphics[width=0.45\columnwidth]{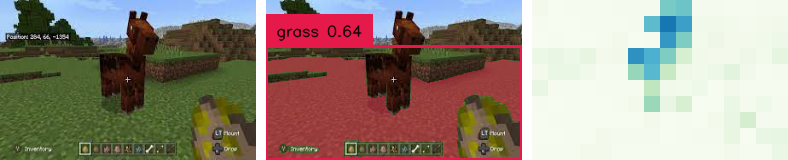}
    \hspace{3mm}
    \includegraphics[width=0.45\columnwidth]{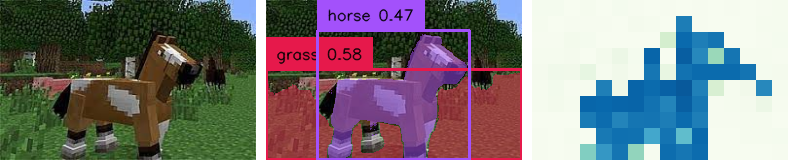}
    \caption{Comparison between Grounded SAM and our method on seven objects.}
    \label{fig:sam}
\end{figure}

We choose one off-the-shelf object detection model, Grounded SAM \cite{liu2023grounding,kirillov2023segany}, to evaluate its effectiveness in Minecraft. In order to conduct a fair side-by-side comparison between it and our method, we Google searched ``minecraft [object name] screenshot'' in the image tab, and selected the first two images that include objects and have them fully in the field of view. The evaluation objects includes \texttt{pig}, \texttt{cow}, \texttt{sheep}, \texttt{mushroom cow}, \texttt{tree}, \texttt{flower}, and \texttt{horse}. We follow the setting in the official demo\footnote{\url{https://github.com/IDEA-Research/Grounded-Segment-Anything/blob/main/grounded_sam_colab_demo.ipynb}} to evaluate the effectiveness of Grounded SAM on detecting these objects in Minecraft. For both Grounded SAM and our modified MineCLIP in this evaluation, we use the same word list which consists of the seven evaluation objects and \texttt{grass}.

The detection results of the two methods are illustrated in \Cref{fig:sam}. For a more detailed evaluation, we quantify the number of objects present in each image, the number detected by Grounded SAM, and the number detected by our method. These quantitative results are summarized in \Cref{tab:sam}. Across all images, there are 24 target objects. Grounded SAM can successfully identify 14 objects, which translates to a detection rate of 58.3\%. In contrast, our method demonstrates a significantly higher efficacy, successfully detecting 22 of the 24 objects, achieving a detection rate of 91.7\%. There are two failures in our method. One is the sunflower in the bottom-right corner of the first flower image, and the other is the tree in the left of the second tree image. In both cases, our method generates some activation in the target patches, but it does not cover the entire object (flower) or is relatively weak (tree). We regard them as failures for a more strict result.

\begin{table}[t]
    \centering
    \caption{Result statistics of Grounded SAM and our method on seven objects.}
    \renewcommand{\arraystretch}{1.2}
    \begin{threeparttable}
    \setlength{\tabcolsep}{5pt}
    \begin{tabular}{cccc}
        \toprule
        Object & Ground-truth & Grounded SAM & Ours \\
        \midrule
        \texttt{pig} & 3 & 3 & 3 \\
        \texttt{cow} & 3 & 2 & 3 \\
        \texttt{sheep} & 2 & 0 & 2 \\
        \texttt{mushroom cow} & 4 & 0 & 4 \\
        \texttt{tree} & 6 & 6 & 5 \\
        \texttt{flower}\tnote{\dag} & 4 & 2 & 3 \\
        \texttt{horse} & 2 & 1 & 2 \\
        \textbf{total} & \textbf{24} & \textbf{14} & \textbf{22} \\
        \bottomrule
    \end{tabular}
    \begin{tablenotes}
        \item[\dag] We also count the two flowers held in players' hands.
    \end{tablenotes}
    \end{threeparttable}
    \label{tab:sam}
\end{table}

\section{Policy Learning Details}

\subsection{Observation Space and Action Space}
\label{app:space}

The observation space adopted in our experiments consists of RGB, compass, GPS, voxels, and biome index. The shape and description of each modality are listed in \Cref{tab:ob_space}. We simplify the original action space of MineDojo \cite{fan2022minedojo} into a 2-dimensional multi-discrete action space. The first dimension contains 12 discrete actions about movement: \textit{no\_op}, \textit{move forward}, \textit{move backward}, \textit{move left}, \textit{move right}, \textit{jump}, \textit{sneak}, \textit{sprint}, \textit{camera pitch} -30, \textit{camera pitch} +30, \textit{camera yaw} -30, and \textit{camera yaw} +30. The second dimension includes 3 discrete actions about interacting with items: \textit{no\_op}, \textit{attack}, and \textit{use}.

\begin{table}[htb]
    \centering
    \caption{Observation space adopted in our experiments.}
    \renewcommand{\arraystretch}{1.2}
    \setlength{\tabcolsep}{5pt}
    \begin{tabular}{cll}
        \toprule
        Modality & Shape & Description \\
        \midrule
        RGB & (3, 160, 256) & Ego-Centric RGB frames. \\
        Compass & (4,) & Sine and cosine of yaw and pitch. \\
        GPS & (3,) & GPS location of the agent. \\
        Voxels & (27,) & Indices of $3\times3\times3$ surrounding blocks. \\
        Biome\_ID & (1,) & Index of the environmental biome. \\
        \bottomrule
    \end{tabular}
    \label{tab:ob_space}
\end{table}

\subsection{Network Architecture}
\label{app:net}

\begin{figure}[t]
    \centering
    \includegraphics[width=0.9\columnwidth]{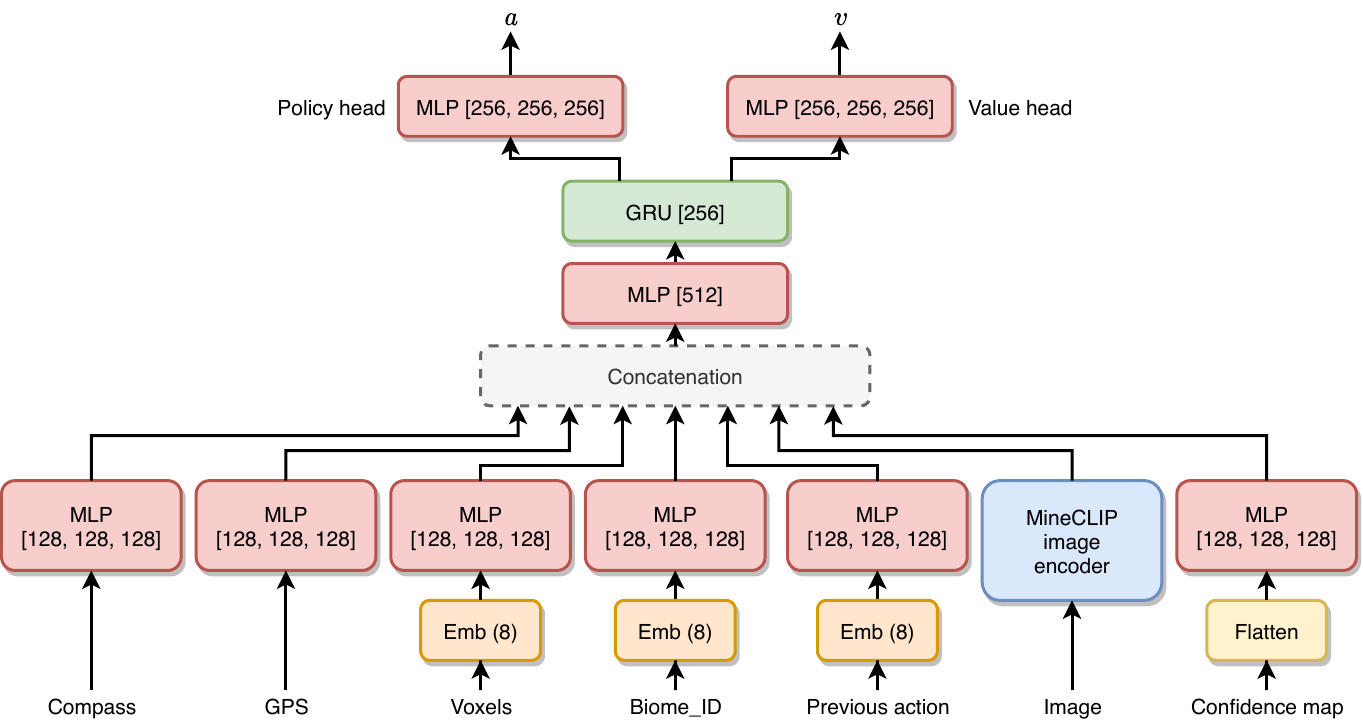}
    \caption{Network architecture of COPL agent.}
    \label{fig:network}
\end{figure}

The input of COPL agent includes observations from the environment listed in \Cref{tab:ob_space}, the agent's action taken at the last time step $a_{t-1}$, and the confidence map. As illustrated in \Cref{fig:network}, all inputs except the RGB image are processed by MLPs with three hidden layers and ReLU activation. In this step, voxels, biome index, and previous action are first embedded into dense vectors. The RGB image is processed using the MineCLIP image encoder to generate an embedding. All these processed features are concatenated and processed by an MLP with one hidden layer and ReLU activation. Then a GRU layer is implemented to integrate the historical information. The policy head and the value head take as input the output of GRU and both process it using an MLP with three hidden layers and ReLU activation. The policy head generates the distribution of actions, and the value head outputs the estimated value of the current state. Some variants are as follows: (1) \textbf{Single-task} model: In single-task experiments, the agent \emph{does not} take as input the confidence map; (2) \textbf{LCRL}: The branch of confidence map is replaced by the MineCLIP text encoder processing the target name or the instruction; (3) \textbf{One-Hot}: The branch of confidence map is replaced by an MLP processing the one-hot vector which indicates the index of the current task. The MLP has one hidden layer with size 32 and ReLU activation.

\section{Single-Task Experiments}

\subsection{Settings}
\label{app:env_single}

Our single-task experiments include four tasks: \texttt{hunt a cow}, \texttt{hunt a sheep}, \texttt{hunt a pig}, and \texttt{hunt a chicken}. The parameters we used to make environments in MineDojo are listed in \Cref{tab:env_single}. In all tasks, the agent spawns in \texttt{plains} biome holding a diamond sword. Several animals including the target spawn near the agent. The agent will receive a +100 reward after successfully killing the target animal. Each episode is limited to a maximum of 200 steps. The challenge lies in the fact that animals will flee after being attacked, thus requiring the agent to keep chasing the target and attacking. Killing a cow, sheep, or pig requires at least two attacks, while killing a chicken only requires at least one attack. Although it takes fewer attacks to kill a chicken, aiming at the small size of the chicken poses an additional challenge. For ablation experiments on Gaussian kernel, we double the initial animals and increase the animal spawn range to 10.

\begin{table}[t]
    \centering
    \caption{Single-task settings in our experiments.}
    \renewcommand{\arraystretch}{1.2}
    \begin{threeparttable}
    \setlength{\tabcolsep}{2pt}
    \begin{tabular}{lcccccc}
        \toprule
        \multicolumn{1}{c}{Task} & Target & Initial Animals & Range\tnote{1} & Inventory & Biome & Length\tnote{2} \\
        \midrule
        \texttt{hunt a cow} & cow & cow, sheep, pig & 7 & diamond\_sword & plains & 200 \\
        \texttt{hunt a sheep} & sheep & cow, sheep, pig & 7 & diamond\_sword & plains & 200 \\
        \texttt{hunt a pig} & pig & cow, sheep, pig & 7 & diamond\_sword & plains & 200 \\
        \texttt{hunt a chicken} & chicken & cow, sheep, chicken & 7 & diamond\_sword & plains & 200 \\
        \bottomrule
    \end{tabular}
    \begin{tablenotes}
        \item[1] Range indicates the spawn range of initial animals.
        \item[2] Length indicates the maximum length of one episode.
    \end{tablenotes}
    \end{threeparttable}
    \label{tab:env_single}
    \vspace{4mm}
\end{table}

\subsection{Learning Curves}
\label{app:single_curve}

\begin{figure}[t]
    \centering
    \begin{subfigure}{0.245\textwidth}
        \includegraphics[width=1.0\textwidth]{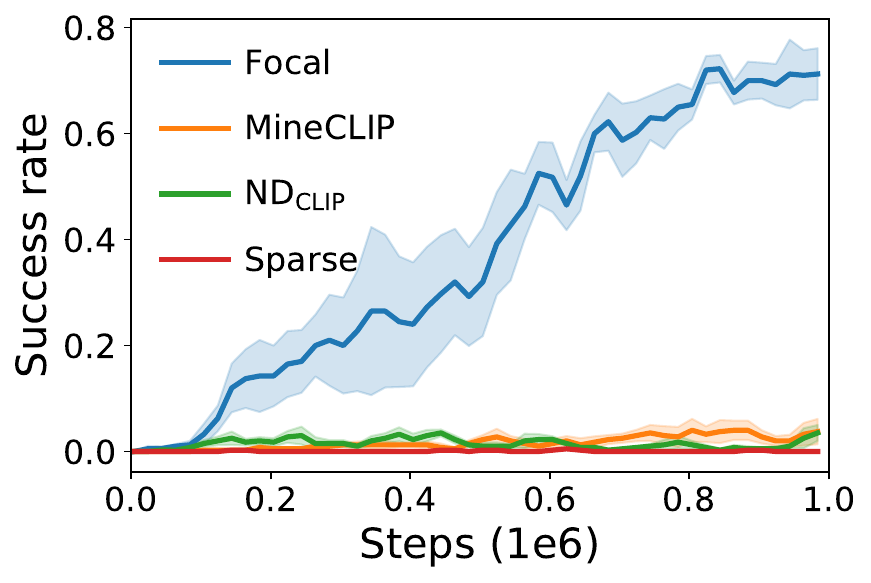}
        \caption{\texttt{hunt a cow}}
    \end{subfigure}
    \hspace{-2mm}
    \begin{subfigure}{0.245\textwidth}
        \includegraphics[width=1.0\textwidth]{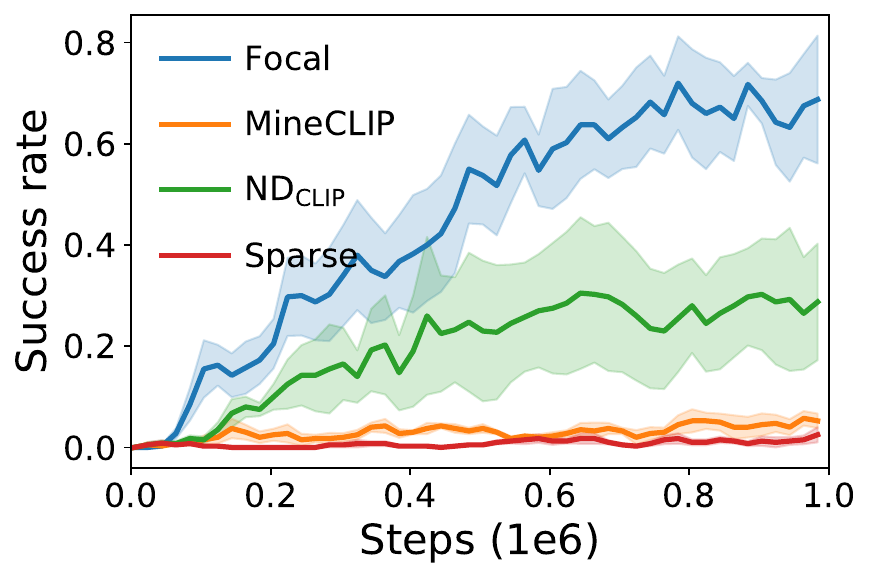}
        \caption{\texttt{hunt a sheep}}
    \end{subfigure}
    \hspace{-2mm}
    \begin{subfigure}{0.245\textwidth}
        \includegraphics[width=1.0\textwidth]{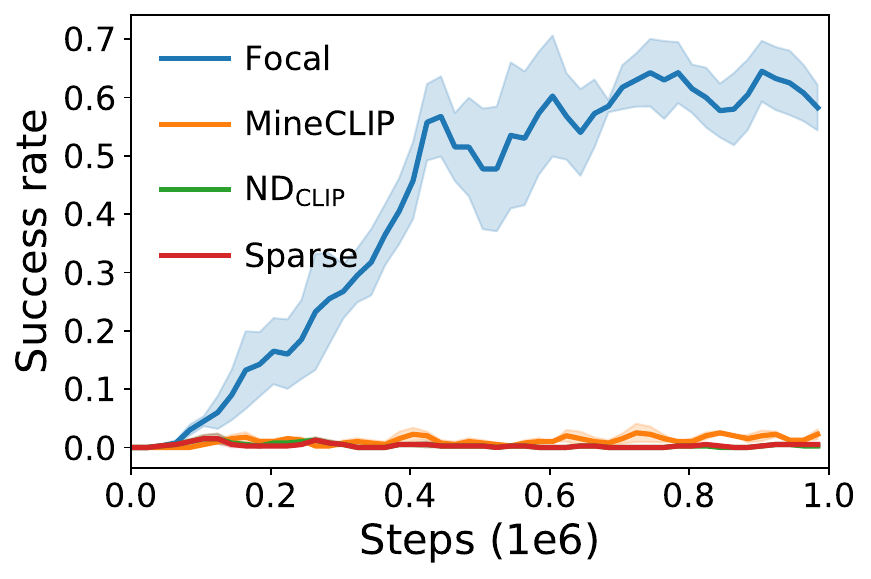}
        \caption{\texttt{hunt a pig}}
    \end{subfigure}
    \hspace{-2mm}
    \begin{subfigure}{0.245\textwidth}
        \includegraphics[width=1.0\textwidth]{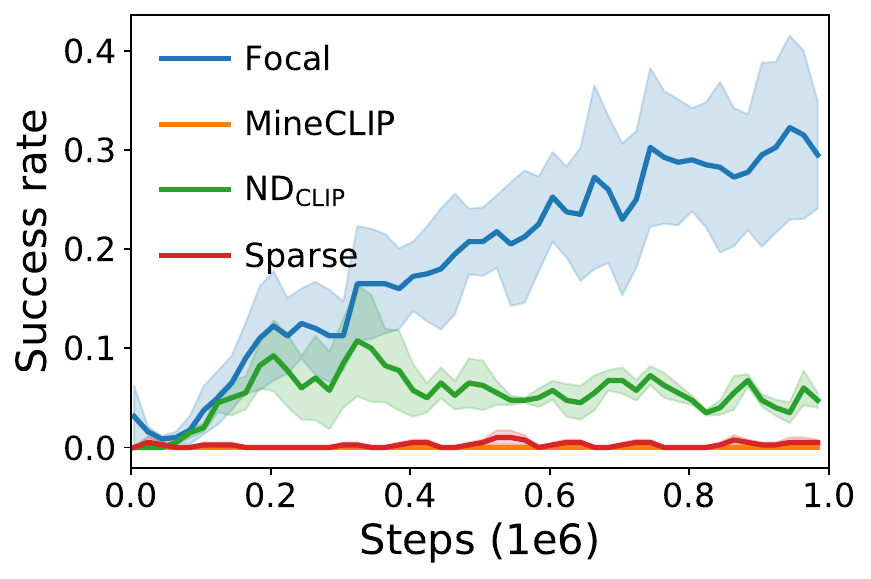}
        \caption{\texttt{hunt a chicken}}
    \end{subfigure}
    \caption{Learning curves of PPO with focal reward, MineCLIP reward, ND\textsubscript{CLIP} reward, and environmental sparse reward only, on four Minecraft tasks.}
    \label{fig:single}
\end{figure}

Learning curves of four methods on for Minecraft tasks are shown in \Cref{fig:single}. Each curve shows the mean success rate of four runs with different seeds and shaded regions indicate standard error (the same applies hereinafter). We can observe that only our focal reward leads to the mastery of all four skills by guiding the agent to consistently approach the target.

\subsection{Additional Experiments}
\label{app:single_more}

\begin{figure}[t]
    \centering
    \begin{subfigure}{0.245\textwidth}
        \includegraphics[width=1.0\textwidth]{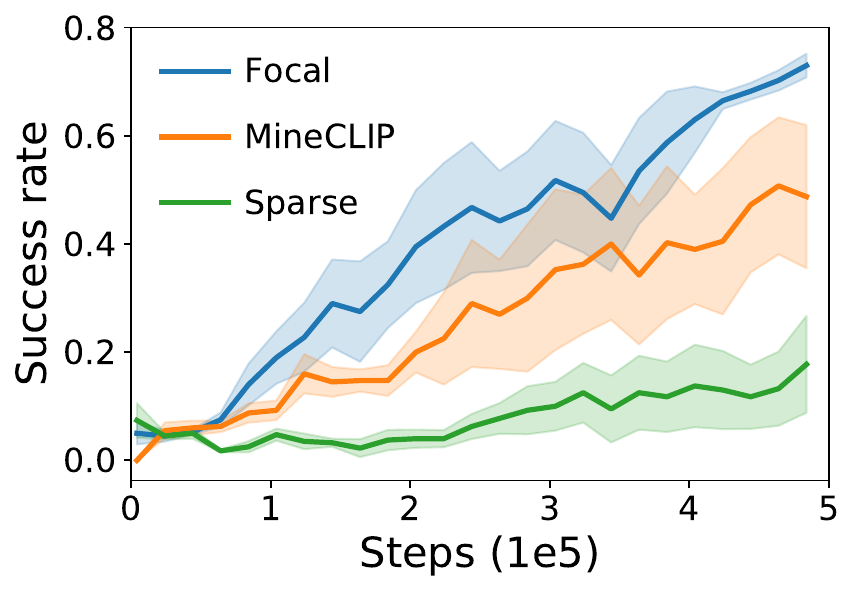}
        \caption{\texttt{milk a cow}}
    \end{subfigure} \quad
    \begin{subfigure}{0.245\textwidth}
        \includegraphics[width=1.0\textwidth]{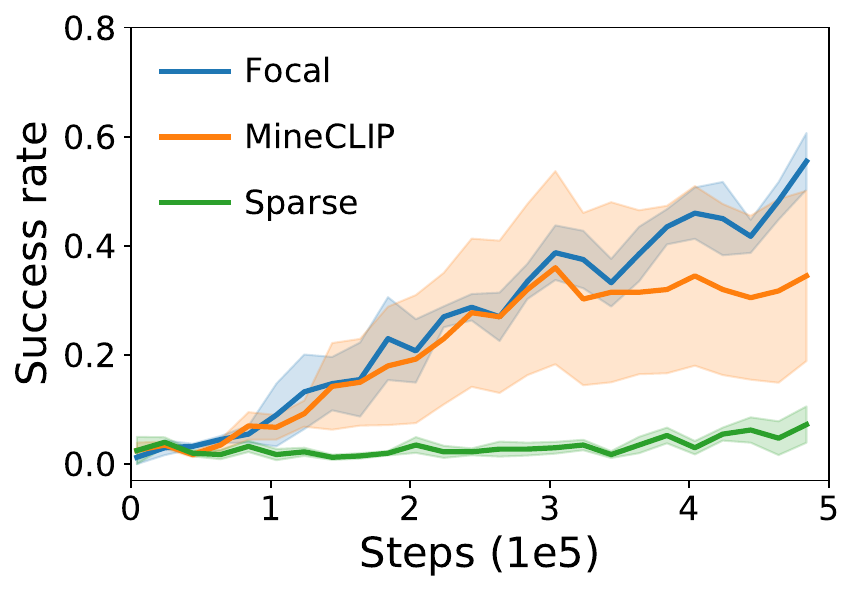}
        \caption{\texttt{milk a cow}}
    \end{subfigure} \quad
    \begin{subfigure}{0.245\textwidth}
        \includegraphics[width=1.0\textwidth]{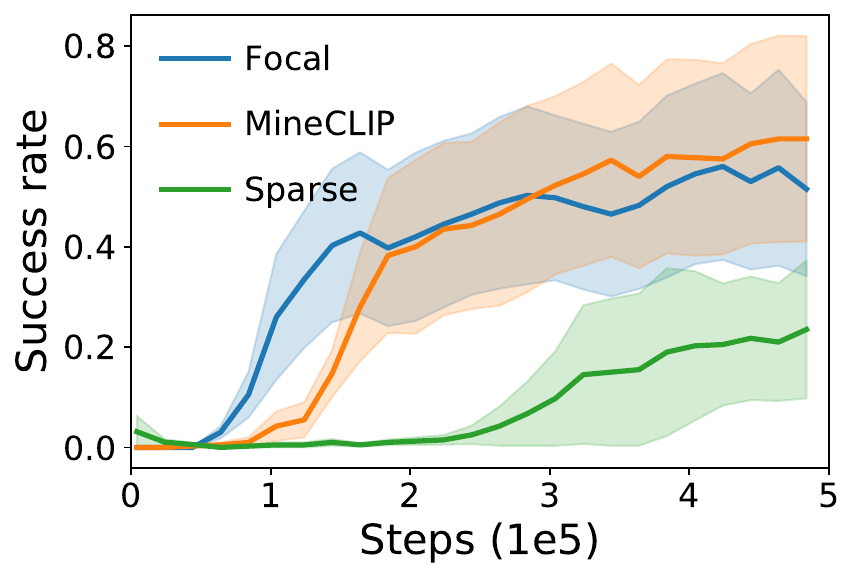}
        \caption{\texttt{milk a cow}}
    \end{subfigure}
    \caption{Learning curves of PPO with focal reward, MineCLIP reward, and environmental sparse reward only, on three Minecraft tasks: (a) \texttt{milk a cow}, (b) \texttt{shear a sheep}, and (c) \texttt{chop a tree}.}
    \label{fig:single_harvest}
\end{figure}

We conduct additional single-task experiments on three harvest tasks including \texttt{milk a cow}, \texttt{shear a sheep}, and \texttt{chop a tree}, where MineCLIP reward achieves nonzero success rates \cite{fan2022minedojo}. The environment parameters for each task can be found in \Cref{tab:env_single_extra}. As shown in \Cref{fig:single_harvest}, our focal reward outperforms MineCLIP reward on \texttt{milk a cow} and \texttt{shear a sheep}. Regarding \texttt{chop a tree}, our focal reward and MineCLIP reward achieve similar performance, both with 3 out of 4 runs having learned this skill. To break a wood block, the agent needs to continuously take \textit{attack} actions for around 6 steps. Therefore, we believe that the main challenge for RL in this task lies in exploration. It is difficult for an RL algorithm, such as PPO, with a stochastic policy to explore and exploit a behavior pattern that requires consecutive actions over 6 steps, especially given the sparse environmental reward signal. Using an off-policy RL algorithm or self-imitation may help address this problem.

\begin{table}[t]
    \centering
    \caption{Single-task settings in additional experiments.}
    \renewcommand{\arraystretch}{1.2}
    \setlength{\tabcolsep}{3pt}
    \begin{tabular}{lcccccc}
        \toprule
        \multicolumn{1}{c}{Task} & Target & Initial Animals & Range & Inventory & Biome & Length \\
        \midrule
        \texttt{milk a cow} & milk\_bucket & cow, sheep, pig & 10 & bucket & plains & 200 \\
        \texttt{shear a sheep} & wool & cow, sheep, pig & 10 & shears & plains & 200 \\
        \texttt{chop a tree} & log & cow, sheep, pig & 7 & golden\_axe & forest & 200 \\
        \bottomrule
    \end{tabular}
    \label{tab:env_single_extra}
    \vspace{4mm}
\end{table}

To quantitatively assess the accuracy of our focal reward as a proxy for estimating the distance between the agent and the target object, we calculate the correlation coefficient between the reward and the ground-truth distance to the target object at each step. The ground-truth distance to the target object is obtained through an internal function of the environment simulator \cite{fan2022minedojo}, which is typically not available during normal gameplay and not available in all environments. Therefore, for generality, we cannot directly use the distance as a reward for training. We run our trained models on four tasks. For each task, we sample 10,000 steps and record focal rewards, MineCLIP rewards, and ground-truth distance. If the target object is not in the sight of the agent, the ground-truth distance is set to the \textit{inf}. Given that the data distribution is not close to a normal distribution, we calculate the Spearman rank correlation coefficient instead of the Pearson correlation coefficient, as shown in \Cref{tab:corr}. The results demonstrate that the correlation between our focal reward and the distance to the target object is higher than the correlation between the MineCLIP reward and the distance, suggesting that our focal reward better approximates the actual distance.

\begin{table}[t]
\caption{Spearman's rank correlation coefficient between rewards and distance to the target object in sight.}
\centering
\renewcommand{\arraystretch}{1.2}
\begin{tabular}{p{3pt}lp{5pt}lp{5pt}lp{5pt}lp{5pt}lp{3pt}}
\toprule
& \multicolumn{1}{c}{\textbf{Rewards}} & & \multicolumn{1}{c}{\texttt{cow}} & & \multicolumn{1}{c}{\texttt{sheep}} & & \multicolumn{1}{c}{\texttt{pig}} & & \multicolumn{1}{c}{\texttt{chicken}} & \\ \midrule
& Focal     & & \textbf{-0.704} & & \textbf{-0.701} & & \textbf{-0.761} & & \textbf{-0.835} & \\
& MineCLIP  & & -0.374 & & -0.585 & & -0.460 & & -0.544 & \\ \bottomrule
\end{tabular}
\label{tab:corr}
\end{table}

\section{Multi-Task Experiments}

\subsection{Settings}
\label{app:env_multi}

\textbf{Hunting domain.} The hunting domain consists of four instructions: ``\texttt{hunt a cow}'', ``\texttt{hunt a sheep}'', ``\texttt{hunt a pig}'', and ``\texttt{hunt a chicken}''. At the start of each episode, one instruction is randomly selected, and an environment is built with the parameters listed in \Cref{tab:env_hunt}. The agent will receive a +100 reward after successfully killing the target animal specified in the instruction. If the agent mistakenly kills the animal which is the target of other instructions, no reward is given and the episode ends. This setup encourages the agent to attack the correct animal rather than indiscriminately attacking any animal.

The object-level generalization evaluation for the hunting domain also contains four instructions: ``\texttt{hunt a mushroom cow}'', ``\texttt{hunt a spider}'', ``\texttt{hunt a \\llama}'', and ``\texttt{hunt a horse}''. The environment parameters can be found in \Cref{tab:env_hunt_eval}. We slightly increase the maximum episode length for ``\texttt{hunt a spider}'', ``\texttt{hunt a llama}'', and ``\texttt{hunt a horse}'', given that killing them requires more attacks as a result of their higher health compared to other animals. For each instruction, we run the test model for 100 episodes to calculate its success rate and precision (same in the harvest domain).

\vspace{2mm}
\noindent \textbf{Harvest domain.} The harvest domain consists of four instructions: ``\texttt{milk a cow}'', ``\texttt{shear a sheep}'', ``\texttt{harvest a flower}'', and ``\texttt{harvest leaves}''. Same as the hunting domain, one instruction is randomly selected at the start of each episode, and an environment is generated with the parameters listed in \Cref{tab:env_harvest}. The agent will receive a +100 reward after successfully acquiring the target item. If the agent mistakenly acquires the target item corresponding to other instructions, no reward is given and the episode ends. Note that \textit{the target item required to finish the task may not always be the same as the target object that the agent needs to approach}. For example, in the instruction ``\texttt{milk a cow}'', the target item is a milk\_bucket, while the target object that the agent needs to approach is a cow. 

\begin{table}[tb]
    \centering
    \caption{Multi-task settings in the hunting domain.}
    \renewcommand{\arraystretch}{1.2}
    \setlength{\tabcolsep}{3pt}
    \begin{tabular}{lcp{0.18\columnwidth}cccc}
        \toprule
        \multicolumn{1}{c}{Instruction} & Target & Initial Animals & Range & Inventory & Biome & Length \\
        \midrule
        \texttt{hunt a cow} & cow & cow, sheep, pig, chicken & 10 & diamond\_sword & plains & 500 \\
        \texttt{hunt a sheep} & sheep & cow, sheep, pig, chicken & 10 & diamond\_sword & plains & 500 \\
        \texttt{hunt a pig} & pig & cow, sheep, pig, chicken & 10 & diamond\_sword & plains & 500 \\
        \texttt{hunt a chicken} & chicken & cow, sheep, pig, chicken & 10 & diamond\_sword & plains & 500 \\
        \bottomrule
    \end{tabular}
    \label{tab:env_hunt}
\end{table}

\begin{table}[tb]
    \centering
    \caption{Generalization evaluation settings in the hunting domain.}
    \renewcommand{\arraystretch}{1.3}
    \begin{threeparttable}
    \begin{footnotesize}
    \arraybackslash
    \begin{tabular}{>{}p{0.18\columnwidth} >{\centering}p{0.15\columnwidth} >{\raggedright}p{0.19\columnwidth}cccc}
        \toprule
        \multicolumn{1}{c}{Instruction} & Target & Initial Animals & Range & Inventory & Biome & Length \\
        \midrule
        \texttt{hunt a mush- room cow} & mushroom cow & mushroom cow, spider, llama, horse & 10 & diamond\_sword & plains & 500 \\
        \texttt{hunt a spider} & spider & mushroom cow, spider, llama, horse & 10 & diamond\_sword & plains & 800 \\
        \texttt{hunt a llama} & llama & mushroom cow, spider, llama, horse & 10 & diamond\_sword & plains & 800 \\
        \texttt{hunt a horse} & horse & mushroom cow, spider, llama, horse & 10 & diamond\_sword & plains & 800 \\
        \bottomrule
    \end{tabular}
    \end{footnotesize}
    \end{threeparttable}
    \label{tab:env_hunt_eval}
\end{table}

\begin{table}[tb]
    \centering
    \caption{Multi-task settings in the harvest domain.}
    \renewcommand{\arraystretch}{1.2}
    \begin{threeparttable}
    \begin{tabular}{>{}p{0.2\columnwidth} cccccc}
        \toprule
        \multicolumn{1}{c}{Instruction} & Target\tnote{1} & Initial Animals & Range & Inventory & Biome & Length \\
        \midrule
        \texttt{milk a cow} & milk\_bucket & cow, sheep, pig & 10 & bucket & plains & 200 \\
        \texttt{shear a sheep} & wool & cow, sheep, pig & 10 & shears & plains & 200 \\
        \texttt{harvest a flo- wer} & red\_flower & cow, sheep, pig & 10 & - & flower\_forest & 200 \\
        \texttt{harvest leaves} & leaves & cow, sheep, pig & 10 & shears & flower\_forest & 200 \\
        \bottomrule
    \end{tabular}
    \begin{tablenotes}
        \item[1] Target here represents the parameter for making a MineDojo environment, \textit{i.e.}, the target item required to finish the task. It differs from the target object specified in the instruction.
    \end{tablenotes}
    \end{threeparttable}
    \label{tab:env_harvest}
\end{table}

The object-level generalization evaluation for the harvest domain contains four instructions: ``\texttt{harvest water}'', ``\texttt{shear a mushroom cow}'', ``\texttt{collect sand}'', and ``\texttt{collect dirt}''. For each instruction, there is a distraction item. If the agent mistakenly acquires the distraction item, no reward is given and the episode ends. For ``\texttt{harvest water}'', the distraction item is milk\_bucket, as the agent can also acquire a bucket of milk with the given empty bucket from a cow. Similarly, the distraction items for the other three instructions are wool, dirt, and sand, respectively.

\begin{table}[tb]
    \centering
    \caption{Generalization evaluation settings in the harvest domain.}
    \renewcommand{\arraystretch}{1.2}
    \arraybackslash
    \begin{tabular}{>{}p{0.2\columnwidth} >{\centering}p{0.12\columnwidth} >{\raggedright}p{0.19\columnwidth}cccc}
        \toprule
        \multicolumn{1}{c}{Instruction} & Target & Initial Animals & Range & Inventory & Biome & Length \\
        \midrule
        \texttt{harvest water} & water\_ bucket & cow, sheep, mushroom cow & 10 & bucket & river & 200 \\
        \texttt{shear a mushroom cow} & mushroom & cow, sheep, mushroom cow & 10 & shears & plains & 200 \\
        \texttt{collect sand} & sand & cow, sheep, mushroom cow & 10 & diamond\_shovel & river & 200 \\
        \texttt{collect dirt} & dirt & cow, sheep, mushroom cow & 10 & diamond\_shovel & river & 200 \\
        \bottomrule
    \end{tabular}
    \label{tab:env_harvest_eval}
\end{table}

Here we briefly introduce the behavior patterns required by the harvest domain instructions. ``\texttt{Milk a cow}'' and ``\texttt{harvest water}'' require the agent to approach the target object (cow/water), aim at it, and take \textit{use} action. ``\texttt{Harvest a flower}'', ``\texttt{collect sand}'', and ``\texttt{collect dirt}'' require the agent to approach the target object (flower/sand/dirt), aim at it, take \textit{attack} action to break it, and move closer to pick up the dropped item. ``\texttt{Shear a sheep}'' and ``\texttt{harvest leaves}'' are the same except that they require taking \textit{use} action instead of \textit{attack} action. In all these tasks except ``\texttt{collect sand}'' and ``\texttt{collect dirt}'', the agent needs only one \textit{attack} or \textit{use} action to finish the task. Therefore, individually, the harvest tasks are easier than the hunting tasks. ``\texttt{Collect sand}'' and ``\texttt{collect dirt}'' require the agent to continuously \textit{attack} multiple times to break a sand or dirt block.

\subsection{Learning Curves}
\label{app:mt_rl}

\begin{figure}[tb]
    \centering
    \begin{subfigure}{0.245\textwidth}
        \includegraphics[width=1.0\textwidth]{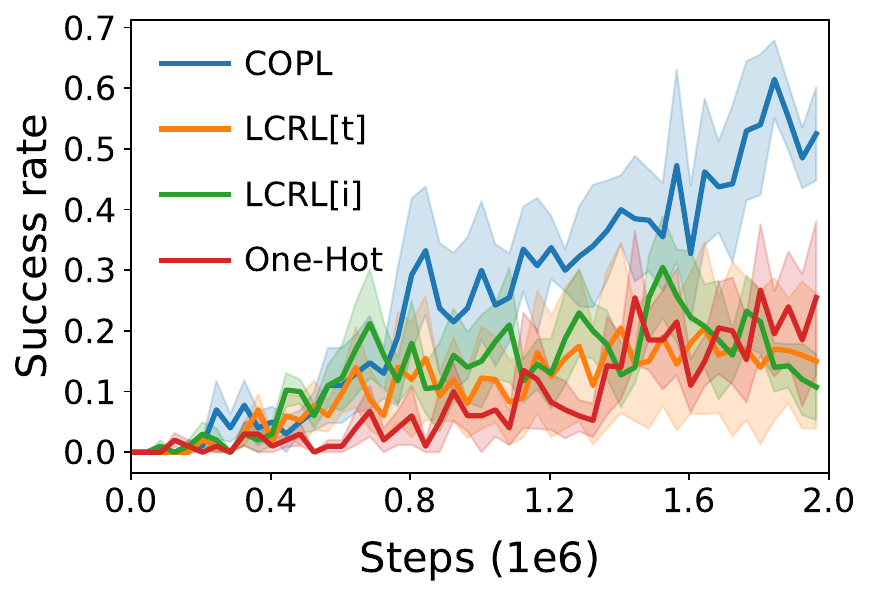}
        \caption{\texttt{hunt a cow}}
    \end{subfigure}
    \hspace{-2mm}
    \begin{subfigure}{0.245\textwidth}
        \includegraphics[width=1.0\textwidth]{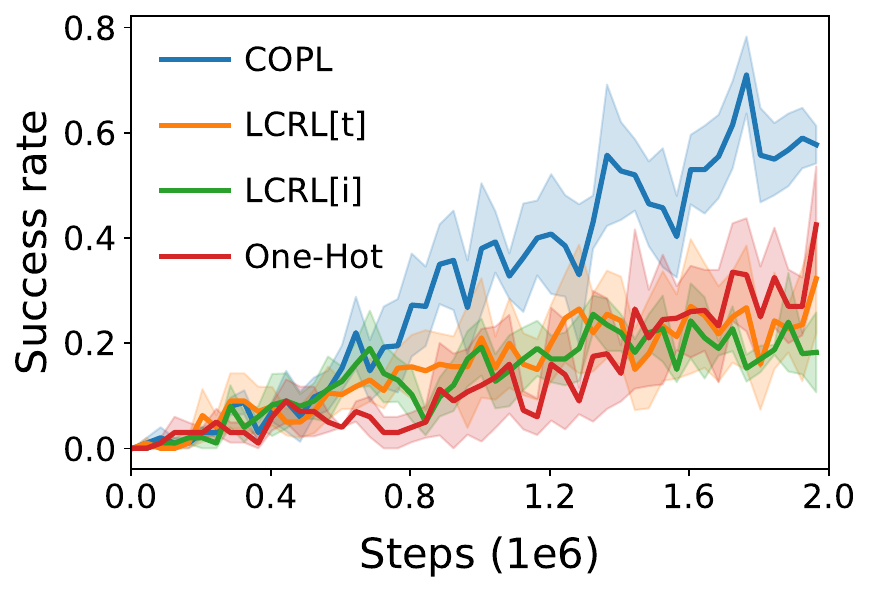}
        \caption{\texttt{hunt a sheep}}
    \end{subfigure}
    \hspace{-2mm}
    \begin{subfigure}{0.245\textwidth}
        \includegraphics[width=1.0\textwidth]{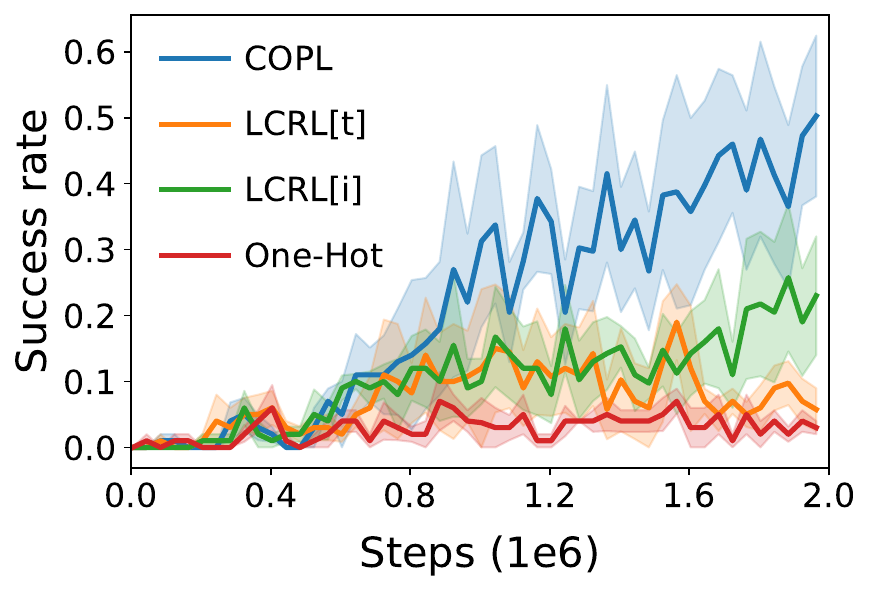}
        \caption{\texttt{hunt a pig}}
    \end{subfigure}
    \hspace{-2mm}
    \begin{subfigure}{0.245\textwidth}
        \includegraphics[width=1.0\textwidth]{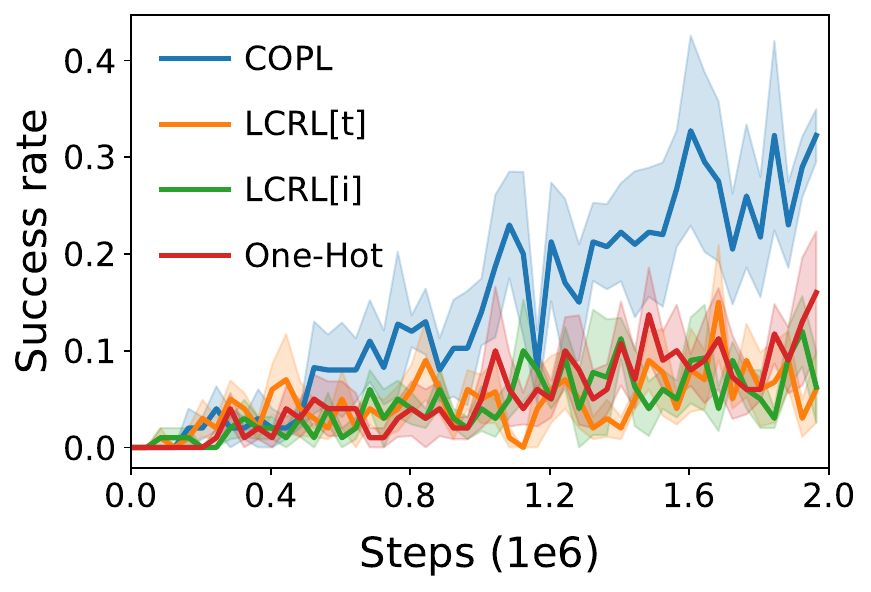}
        \caption{\texttt{hunt a chicken}}
    \end{subfigure}
    \caption{Learning curves of COPL, LCRL[t], LCRL[i], and One-Hot on four hunting instructions: (a) ``\texttt{hunt a cow}'', (b) ``\texttt{hunt a sheep}'', (c) ``\texttt{hunt a pig}'', and (d)``\texttt{hunt a chicken}''.}
    \label{fig:mt_combat_each}
\end{figure}

\begin{figure}[tb]
    \centering
    \begin{subfigure}{0.245\textwidth}
        \includegraphics[width=1.0\textwidth]{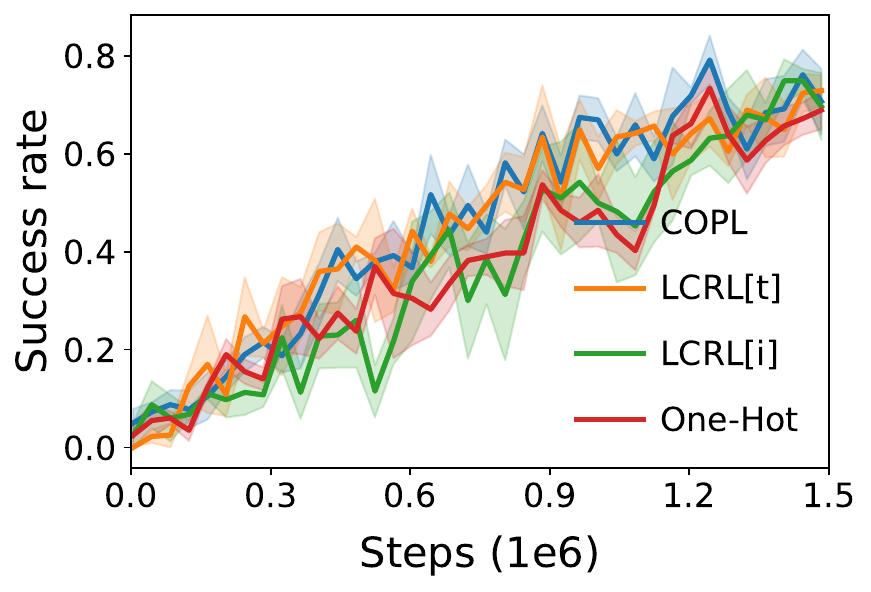}
        \caption{\texttt{milk a cow}}
    \end{subfigure}
    \hspace{-2mm}
    \begin{subfigure}{0.245\textwidth}
        \includegraphics[width=1.0\textwidth]{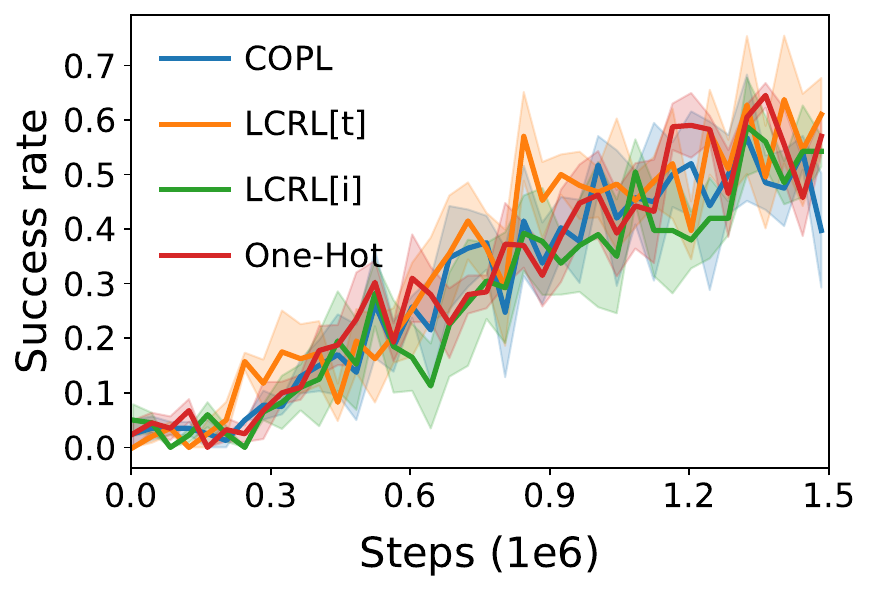}
        \caption{\texttt{shear a sheep}}
    \end{subfigure}
    \hspace{-2mm}
    \begin{subfigure}{0.245\textwidth}
        \includegraphics[width=1.0\textwidth]{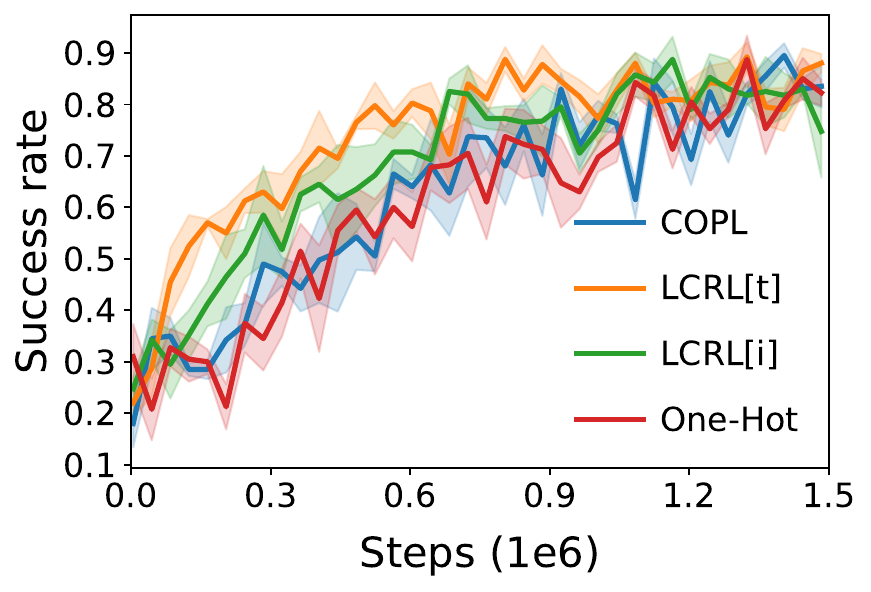}
        \caption{\texttt{harvest a flower}}
    \end{subfigure}
    \hspace{-2mm}
    \begin{subfigure}{0.245\textwidth}
        \includegraphics[width=1.0\textwidth]{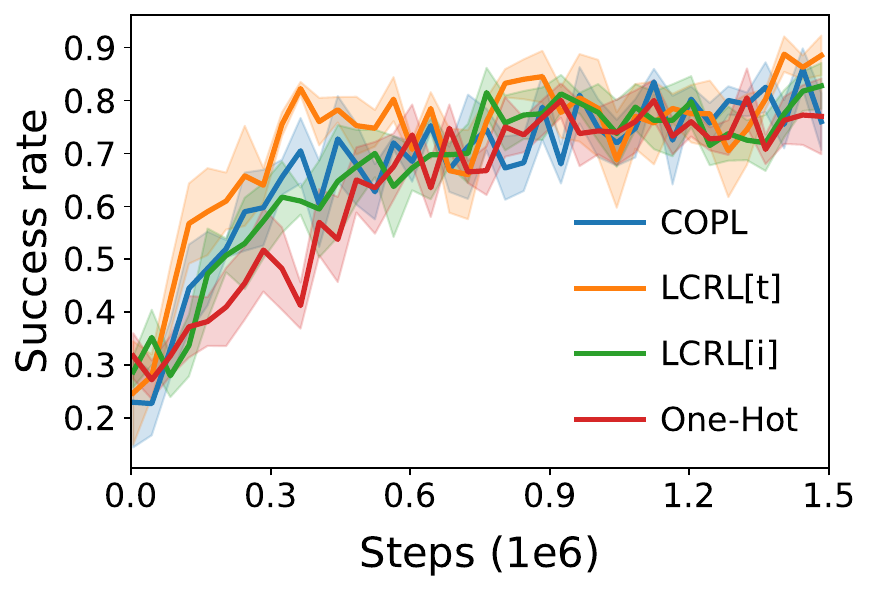}
        \caption{\texttt{harvest leaves}}
    \end{subfigure}
    \caption{Learning curves of COPL, LCRL[t], LCRL[i], and One-Hot on four harvest instructions: (a) ``\texttt{milk a cow}'', (b) ``\texttt{shear a sheep}'', (c) ``\texttt{harvest a flower}'', and (d)``\texttt{harvest leaves}''.}
    \label{fig:mt_harvest_each}
\end{figure}

We show the learning curves of COPL, LCRL[t], LCRL[i], and One-Hot on each task. As illustrated in \Cref{fig:mt_combat_each,fig:mt_harvest_each}.

\subsection{Precision}
\label{app:prcsn}

In the hunting domain, precision is defined as the number of correct kills on the specified target animal divided by the number of kills on any animal. The high precision, as reported in \Cref{tab:mt_combat_prcsn}, proves COPL's ability to distinguish the target animal from other animals, rather than indiscriminately attacking them, even if these animals are all out of the training scope. As for the harvest domain, precision is related to the distraction item claimed in \Cref{app:env_multi}, as harvesting the distraction item requires the same tool as the target item. Precision shown in \Cref{tab:mt_harvest_prcsn} is defined as the number of times correctly harvesting the specified target item divided by the total number of times harvesting the target item or the distraction item. Similar to the results in the hunting domain, COPL exhibits better identification ability on unseen targets, compared to LCRL[t] and LCRL[i], suggesting that the zero-shot object-level generalization of COPL emerges from grounding the unseen target object in a simple two-dimensional visual representation, given that all methods perform equally on training tasks.

\begin{table}[t]
    \centering
    \renewcommand{\arraystretch}{1.2}
    \caption{Precision (\%) on hunting test tasks.}
    \begin{tabular}{p{1pt}lp{3pt}lp{3pt}lp{3pt}lp{3pt}lp{3pt}lp{1pt}} \toprule
     & \multicolumn{1}{c}{\textbf{Tasks}} & & \multicolumn{1}{c}{COPL} &  & \multicolumn{1}{c}{LCRL[t]} & & \multicolumn{1}{c}{LCRL[i]} & & \multicolumn{1}{c}{\cite{cai2023open}} & & \multicolumn{1}{c}{STEVE-1} & \\ \midrule
     & \texttt{llama}   & & \textbf{88.8±7.5}   & & 43.3±13.5   & & 55.8±16.7   & & 36.7    & & 6.3 & \\
     & \texttt{horse}   & & \textbf{89.8±7.5}   & & 13.3±2.2    & & 14.8±11.2    & & 28.0    & & 9.4 & \\
     & \texttt{spider}  & & \textbf{96.8±1.9}   & & 47.0±33.0   & & 56.5±40.0   & & 23.3    & & 93.6 & \\
     & \texttt{mushroom} & & \textbf{97.0±2.0}  & & 69.0±26.9   & & 0.0±0.0     & & 55.3    & & 6.8  & \\
     & Avg.              & & \textbf{93.1±3.4}  & & 41.5±3.8    & & 42.3±2.1    & & 35.8    & & 29.0 & \\ \bottomrule
    \end{tabular}
\label{tab:mt_combat_prcsn}
\end{table}

\begin{table}[tb]
    \centering
    \renewcommand{\arraystretch}{1.2}
    \caption{Precision (\%) on harvest test tasks.}
    \begin{tabular}{p{1pt}lp{3pt}lp{3pt}lp{3pt}lp{3pt}lp{1pt}} \toprule
     & \multicolumn{1}{c}{\textbf{Tasks}} & & \multicolumn{1}{c}{COPL} &  & \multicolumn{1}{c}{LCRL[t]} & & \multicolumn{1}{c}{LCRL[i]} & & \multicolumn{1}{c}{STEVE-1} & \\ \midrule
     & \texttt{water}    & & 55.8±11.9  & & 31.8±11.6   & & 28.5±3.0   & & \textbf{91.3} & \\
     & \texttt{mushroom} & & \textbf{74.8±10.5}  & & 56.3±8.2   & & 48.8±9.1    & & 0.0 & \\
     & \texttt{sand}     & & \textbf{70.5±31.2}  & & 10.5±12.6  & & 14.3±10.9    & & 12.0 & \\
     & \texttt{dirt}     & & 95.3±2.6  & & 57.8±43.1   & & 60.8±23.5   & & \textbf{98.6}  & \\
     & Avg.              & & \textbf{74.1±10.4}  & & 39.3±16.4   & & 38.1±7.3    & & 50.5 & \\ \bottomrule
    \end{tabular}
\label{tab:mt_harvest_prcsn}
\end{table}

\section{Hyperparameters}
\label{app:param}

\subsection{PPO Hyperparameters}
\label{app:ppo}

In our experiments, we use PPO \cite{schulman2017proximal} as our base RL algorithm. \Cref{tab:ppo} lists the hyperparameters for PPO across all tasks. Unlike MineAgent \cite{fan2022minedojo}, our implementation does not include self-imitation learning and action smoothing loss. We find that vanilla PPO is able to achieve high performance in our experiments. For single-task experiments, we train RL models for 1,000,000 environment steps. For multi-task experiments in the hunting domain, we train RL models for 2,000,000 environment steps. For multi-task experiments in the harvest domain, we train RL models for 1,500,000 environment steps.

\begin{table}[tb]
    \centering
    \caption{Hyperparameters for PPO across all tasks.}
    \renewcommand{\arraystretch}{1.2}
    \begin{tabular}{cc}
        \toprule
        Hyperparameter & Value \\
        \midrule
        num steps & 1000 \\
        num envs & 4 \\
        num minibatches & 4 \\
        num epoches & 8 \\
        GAE lambda & 0.95 \\
        discounted gamma & 0.99 \\
        entropy coef & 0.005 \\
        PPO clip & 0.2 \\
        learning rate & 1e-4 \\
        optimizer & Adam \\
        recurrent data chunk length & 10 \\
        gradient clip norm & 10.0 \\
        network initialization & orthogonal \\
        normalize advantage & true \\
        \bottomrule
    \end{tabular}
    \label{tab:ppo}
    \vspace{4mm}
\end{table}

\subsection{Intrinsic Reward Coefficient}
\label{app:coef}

\begin{figure}[t]
    \centering
    \begin{subfigure}{0.245\textwidth}
        \includegraphics[width=1.0\textwidth]{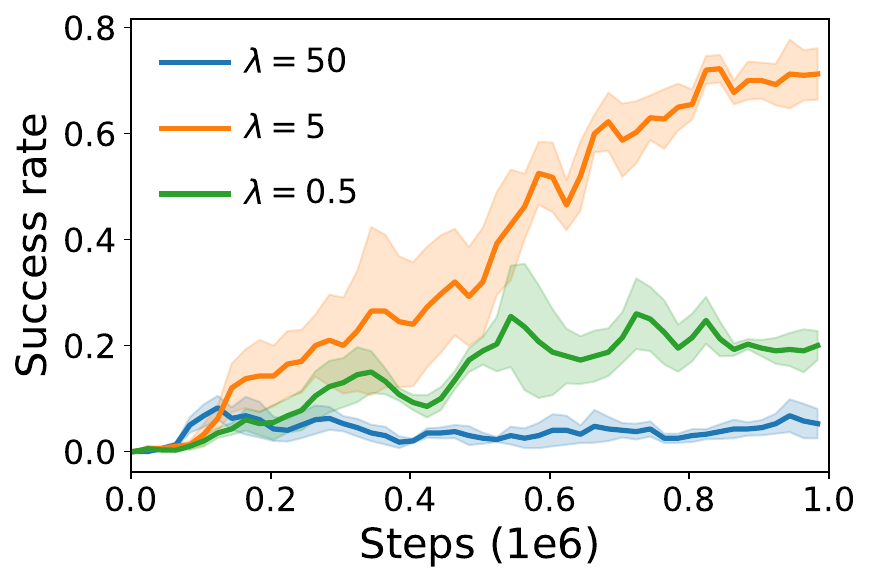}
        \caption{\texttt{hunt a cow}}
        \label{fig:lambda_cow}
    \end{subfigure}
    \hspace{-2mm}
    \begin{subfigure}{0.245\textwidth}
        \includegraphics[width=1.0\textwidth]{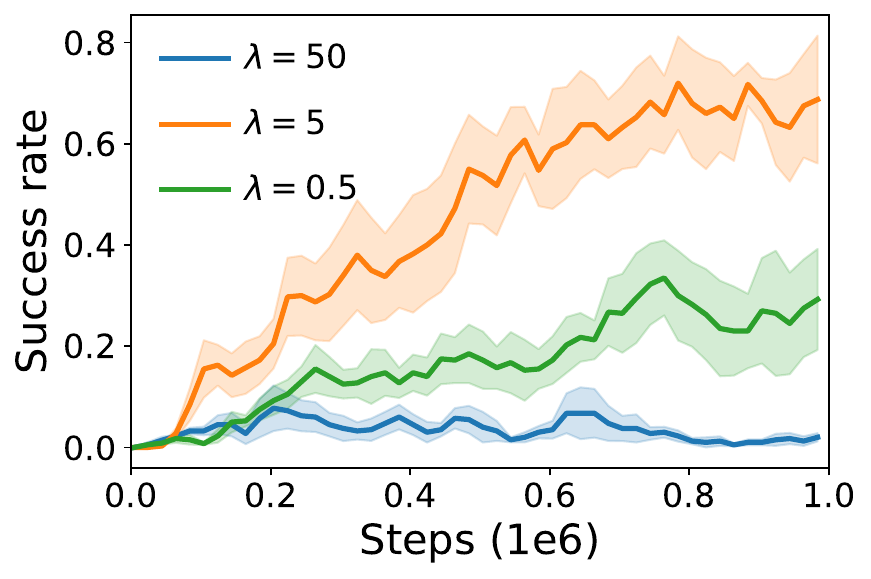}
        \caption{\texttt{hunt a sheep}}
        \label{fig:lambda_sheep}
    \end{subfigure}
    \hspace{-2mm}
    \begin{subfigure}{0.245\textwidth}
        \includegraphics[width=1.0\textwidth]{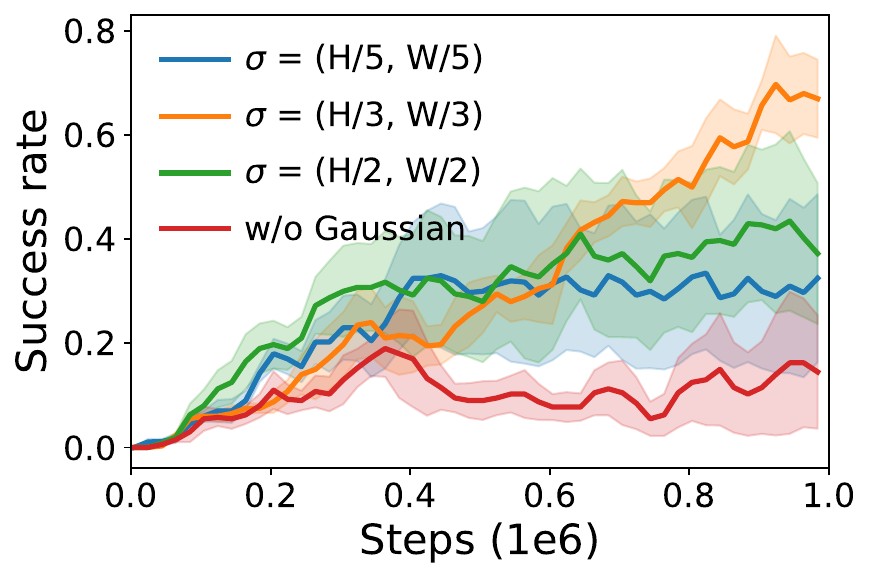}
        \caption{\texttt{hunt a cow}}
        \label{fig:kernel_cow}
    \end{subfigure}
    \hspace{-2mm}
    \begin{subfigure}{0.245\textwidth}
        \includegraphics[width=1.0\textwidth]{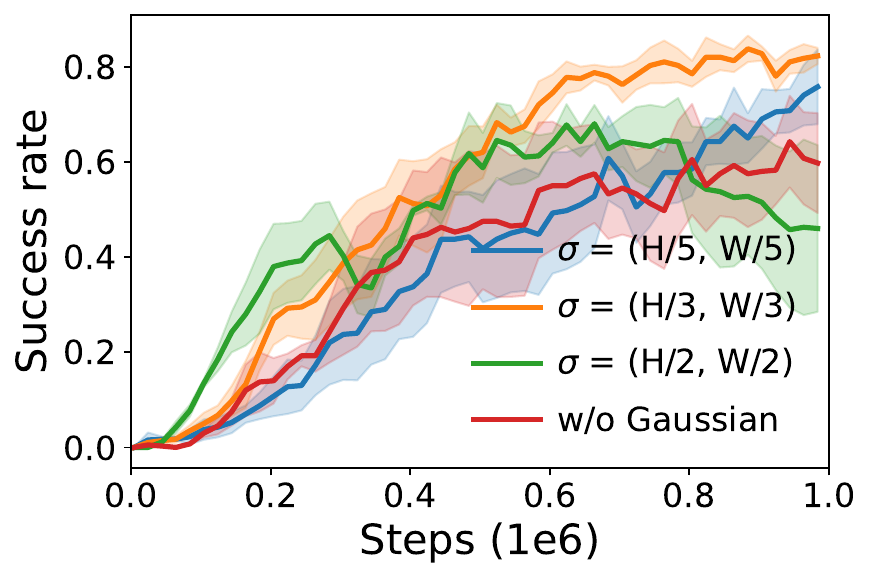}
        \caption{\texttt{hunt a sheep}}
        \label{fig:kernel_sheep}
    \end{subfigure}
    \caption{(a)(b) Learning curves of PPO using the focal reward with different $\lambda$ on two Minecraft tasks. (c)(d) Learning curves of PPO using the focal reward with different Gaussian variances on two Minecraft tasks.}
\end{figure}

To determine the optimal scale of intrinsic reward that can effectively guide reinforcement learning while avoiding conflicts with the environmental reward, we conduct an experiment to evaluate the performance of our focal reward with different $\lambda$ values. \Cref{fig:lambda_cow,fig:lambda_sheep} illustrates the performance of our focal reward with different $\lambda$, including 0.5, 5, and 50, on \texttt{hunt a cow} and \texttt{hunt a sheep}. Focal reward with $\lambda=5$ outperforms $\lambda=50$ and $\lambda=0.5$ on two tasks. Therefore, we consistently set $\lambda=5$ for all experiments in the main text. 

Regarding the MineCLIP reward, we set the coefficient to 1.0, following the original setting of MineAgent in \cite{fan2022minedojo}. The optimal coefficient of ND reward in \cite{tam2022semantic} for \textit{find} task is 0.003, and its sparse environmental reward is 1.0. Considering the environmental reward is set to 100 in our experiments, we decided to increase the coefficient for ND\textsubscript{CLIP} from 0.003 to 0.3 in our implementation.

\subsection{Gaussian Kernel}
\label{app:gauss}

The introduction of a Gaussian kernel is to guide the agent to center a target object within its field of view. The Gaussian kernel should create a high contrast between the center and the edge, as well as between the edge and areas outside the field of view. Therefore, the variance of the Gaussian kernel would influence the performance of the focal reward. To evaluate the impact of different variances, we conduct an experiment with $\sigma=(H/5,W/5)$, $\sigma=(H/3,W/3)$, and $\sigma=(H/2,W/2)$. As illustrated in \Cref{fig:kernel_cow,fig:kernel_sheep}, $\sigma=(H/3,W/3)$ outperforms the others. We suppose that a wider Gaussian kernel with $\sigma=(H/2,W/2)$ fails to provide sufficient contrast between the center and the edge. Conversely, a narrower Gaussian kernel with $\sigma=(H/5,W/5)$ cannot provide sufficient contrast between the edge and areas outside the field of view.

\section{Baselines Implementation}
\label{app:baseline}

\textbf{MineCLIP.} We adopt the provided prompt templates in MineDojo to design task prompts for MineCLIP reward computation in single-task experiments. For hunting tasks, we use the prompt ``\texttt{hunt a \{animal\} on plains with a diamond sword}''. For additional harvest tasks in \Cref{app:single_more}, we use the prompts ``\texttt{obtain milk from a cow in plains with an empty bucket}'', ``\texttt{shear a sheep in plains with shears}'', and ``\texttt{chop trees to obtain log with a golden axe}'', respectively.

\vspace{2mm}
\noindent \textbf{\cite{cai2023open}.} We use the released plains model\footnote{\url{https://github.com/CraftJarvis/MC-Controller}} for evaluation. The goal is set to be the name of the target animal.

\begin{table}[tb]
    \centering
    \renewcommand{\arraystretch}{1.2}
    \caption{Success rates (\%) of STEVE-1 with different prompts.}
    \begin{tabular}{c >{\centering}p{0.1\columnwidth} >{\centering}p{0.1\columnwidth} c}
    \toprule
     Animals & ``\texttt{kill}'' & ``\texttt{hunt}'' & ``\texttt{combat}'' \\
     \midrule
     \texttt{cow}     & 6  & 0  & 0  \\
     \texttt{sheep}   & 14 & 0  & 6  \\
     \texttt{pig}     & 9  & 0  & 4  \\
     \texttt{chicken} & 6  & 0  & 7  \\
     \bottomrule
    \end{tabular}
    \label{tab:steve-1}
    \vspace{4mm}
\end{table}

\vspace{2mm}
\noindent \textbf{STEVE-1 \cite{lifshitz2023steve}.} We use the released model\footnote{\url{https://github.com/Shalev-Lifshitz/STEVE-1}} for evaluation. However, STEVE-1 is designed for another simulator, MineRL \cite{guss2019minerl}, with a different action space from MineDojo. We build a wrapper to map STEVE-1's actions into the action space of MineDojo. As noted in the STEVE-1 paper, prompt engineering significantly impacts its performance. Therefore, we attempt three templates for the hunting domain tasks, including ``\texttt{kill a \{animal\}}'', ``\texttt{hunt a \{animal\}}'', and ``\texttt{combat a \{animal\}}''. As shown in \Cref{tab:steve-1}, ``\texttt{kill a \{animal\}}" achieves the highest performance and STEVE-1 cannot understand the original instruction ``\texttt{hunt a \{animal\}}'' at all. Consequently, we use ``\texttt{kill a \{animal\}}'' as prompts given to STEVE-1 for the experiments in the main text. For tasks in the harvest domain, we use prompts ``\texttt{milk a cow}'', ``\texttt{shear a sheep}'', ``\texttt{break a flower}'', ``\texttt{break leaves}'', ``\texttt{collect water}'', ``\texttt{shear a mushroom}'', ``\texttt{collect sand}'', and ``\texttt{collect dirt}'', respectively. The verbs \texttt{break} and \texttt{collect} are selected by referring to the prompts provided in the STEVE-1 paper. ``\texttt{Milk a cow}'', ``\texttt{shear a sheep}'', and ``\texttt{shear a mushroom cow}'' follow original instructions, as we find that ``\texttt{collect \{milk/wool/mushroom\}}'' does not work.

\section{Creative Tasks}
\label{app:creative}

\begin{figure}[tb]
    \centering
    \begin{subfigure}{0.23\textwidth}
        \includegraphics[width=1.0\textwidth]{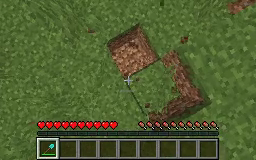}
        \caption{}
        \label{fig:dig_mc_0}
    \end{subfigure}
    \begin{subfigure}{0.23\textwidth}
        \includegraphics[width=1.0\textwidth]{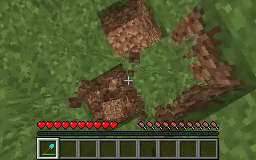}
        \caption{}
        \label{fig:dig_mc_1}
    \end{subfigure}
    \quad
    \begin{subfigure}{0.23\textwidth}
        \includegraphics[width=1.0\textwidth]{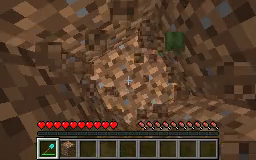}
        \caption{}
        \label{fig:dig_f_0}
    \end{subfigure}
    \begin{subfigure}{0.23\textwidth}
        \includegraphics[width=1.0\textwidth]{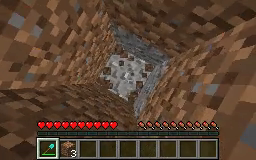}
        \caption{}
        \label{fig:dig_f_1}
    \end{subfigure}
    \caption{(a)(b) Screenshots of the agent trained with MineCLIP reward. (c)(d) Screenshots of the agent trained with our focal reward targeting \texttt{hole}.}
\end{figure}

For \texttt{dig a hole}, the agent spawns with a diamond shovel; for \texttt{lay the carpet}, the agent spawns with 64 carpets. For each task, we train an agent with MineCLIP reward and an agent with our focal reward. The prompts used to calculate MineCLIP reward are ``\texttt{dig a hole}'' and ``\texttt{put carpets on the floor}'', respectively. We run the trained models in the environment and record the agent's depth and the number of placed carpets, averaged on 10 episodes. As illustrated in \Cref{fig:carpet}, the agents trained with MineCLIP reward and our focal reward targeting \texttt{carpet} show the same behavior pattern that keeps laying carpets. However, on \texttt{dig a hole}, the two agents learn different behaviors, as shown in \Cref{fig:dig}: the agent trained with our focal reward targeting \texttt{hole} keeps getting deeper, while the elevation of the one trained with MineCLIP reward does not change too much. By examining their trajectories in the environment, we find that the agent trained with MineCLIP reward tends to dig one block and then stand beside this shallow hole and look at it, as shown in \Cref{fig:dig_mc_0,fig:dig_mc_1}. In contrast, the agent trained with our focal reward stands inside the dug hole and continuously digs downwards, as shown in \Cref{fig:dig_f_0,fig:dig_f_1}. Both behavior patterns are consistent with the description of ``\texttt{dig a hole}'' and can be considered reasonable.

\end{document}